%% file: ICLR_2025_Template/iclr2025_conference.tex
\documentclass{article} 
\usepackage{iclr2025_conference,times}

\input{math_commands.tex}

\usepackage[utf8]{inputenc} 
\usepackage[T1]{fontenc}    
\usepackage{url}            
\usepackage{booktabs}      
\usepackage{amsfonts}       
\usepackage{nicefrac}       
\usepackage{microtype}      
\usepackage[x11names]{xcolor}
\definecolor{CadetBlue}{RGB}{95,158,160}
\usepackage{amssymb}
\usepackage{amsmath}
\usepackage{algorithm}
\usepackage{wrapfig}
\usepackage[utf8]{inputenc} 
\usepackage[T1]{fontenc}    
\usepackage{url}            
\usepackage{booktabs}       
\usepackage{amsfonts}       
\usepackage{graphicx}
\usepackage{nicefrac}       
\usepackage{microtype}      
\usepackage{times}
\usepackage{epsfig}
\usepackage{multirow}
\usepackage{caption}
\usepackage{subcaption}
\usepackage{stfloats}
\usepackage{enumitem}
\usepackage{algpseudocode}
\usepackage{lipsum}
\usepackage{bbm}
\usepackage{wrapfig}\definecolor{mydarkblue}{rgb}{0.082,0.376,0.510}
\usepackage{wrapfig}\definecolor{darkred}{rgb}{0.6,0,0}
\usepackage[colorlinks,
            linkcolor = darkred,
            urlcolor  = darkred, 
            citecolor= mydarkblue,
            ]{hyperref}
\usepackage[labelformat=simple]{subcaption}
\usepackage[most]{tcolorbox}
\usepackage{listings} 
\definecolor{changecolor}{RGB}{0, 0, 0}
\title{Mitigating Mismatch within Reference-based Preference Optimization}

 \iclrfinalcopy

\author{Suqin Yuan\textsuperscript{1} \quad Xingrui Yu\textsuperscript{2} \quad Jiyang Zheng\textsuperscript{1, 3} \\ \textbf{Lei Feng\textsuperscript{4}\footnotemark[1] \quad Dadong Wang\textsuperscript{3} \quad Ivor Tsang\textsuperscript{2, 5} \quad  Tongliang Liu\textsuperscript{1}\thanks{Corresponding authors.}} \\
\textsuperscript{1} Sydney AI Centre, The University of Sydney \quad
\textsuperscript{2} CFAR, A*STAR \\
\textsuperscript{3} CSIRO, Data61 \quad
\textsuperscript{4} Southeast University \quad
\textsuperscript{5} Nanyang Technological University
}

\begin{document}

\maketitle

\begin{abstract}
Direct Preference Optimization (DPO) has become the \textit{de facto} standard for offline preference alignment of large language models, but its reliance on a reference policy introduces a critical tension. 
DPO weighs each update relative to a reference, which stabilizes the training by regularizing the updates within a trusted region. This reliance becomes problematic for pessimistic pairs, where the reference model prefers the rejected response. For these pairs, DPO prematurely attenuates the gradient as soon as the policy margin ($\Delta_\theta$) merely beats the reference margin ($\Delta_{\mathrm{ref}}$) even if the policy is still wrong ($\Delta_{\theta}<0$). We name this failure \emph{premature satisfaction}, which is a concrete form of the \emph{training–inference mismatch}.
Reference-free objectives remove this mismatch by optimizing the absolute margin, but at the cost of discarding the stabilizing signal of the reference. We \textcolor{changecolor}{mitigate} this tension with \emph{Hybrid-DPO (HyPO)}, a drop-in modification to DPO that applies reference \emph{conditionally}: \emph{HyPO} behaves exactly like DPO when the reference is optimistic or neutral, and it treats the reference as neutral when it is pessimistic by replacing $\Delta_\theta-\Delta_{\mathrm{ref}}$ with $\Delta_\theta-\max\{0,\Delta_{\mathrm{ref}}\}$. This one-line change strictly strengthens per-example learning signals on pessimistic pairs while preserving DPO’s objective form and computational cost. By conditionally debiasing the pessimistic reference signal, \emph{HyPO} mitigates \emph{premature satisfaction}; empirically, across preference alignment, \emph{HyPO} improves inference-aligned metrics and achieves higher pairwise win rates. Our results provide evidence that direct preference alignment could be enhanced by conditionally debiasing the reference signal, rather than discarding it.
\end{abstract} 

\section{Introduction}
\label{sec:intro}

Aligning Large Language Models (LLMs) with human preferences and values is paramount for developing safe, helpful, and reliable AI systems \citep{askell2021general}. The pioneering Reinforcement Learning from Human Feedback (RLHF) paradigm \citep{christiano2017deep} achieved this by first training a reward model on human preference pairs \citep{ziegler2019fine, stiennon2020learning, ouyang2022training, bradley1952rank, schulman2017proximal, bradley1952rank}, and then using reinforcement learning \citep{schulman2017proximal} to optimize the LLM for reward maximization, usually with a KL-divergence penalty to regularize deviation from a reference policy \citep{ouyang2022training}. Despite the widespread success, RLHF's multi-stage process suffers from complexity, instability, and high computational cost \citep{sharma2024critical, gao2023scaling}. Addressing these challenges, Direct Preference Optimization (DPO) \citep{rafailov2023direct} emerged as a compelling alternative, offering a significant simplification by analytically deriving a direct policy loss. This loss allows for direct optimization on pairwise preference data $(x, y_w, y_l)$, bypassing explicit reward modeling. Its mechanism implicitly enforces KL regularization by comparing the policy to a reference model ($\pi_{\text{ref}}$), making the loss dependent on the log-likelihood margin $\Delta_\theta - \Delta_{\text{ref}}$. DPO has gained widespread adoption due to its stability, efficiency, and strong empirical performance \citep{rafailov2023direct, d2024anchored, zhao2023slic, yuan2023rrhf, rashidinejadsail}.

While DPO is simple and effective, its reliance on a reference model introduces a tension that has led to two seemingly opposing lines of work. One argues that the reference should be removed due to a \emph{training–inference mismatch} \citep{chen2024preference, meng2024simpo}: DPO optimizes a reference-relative margin (e.g., $\Delta_\theta - \Delta_{\mathrm{ref}}$), whereas inference depends on \emph{absolute} log-likelihoods with no reference. Empirically, after DPO alignment, the agreement between the implicit-reward ordering and the likelihood ordering is only $\sim$50\%, motivating reference-free objectives that directly optimize the inference metric and often outperform DPO \citep{meng2024simpo}. The other line takes the opposite view: the reference can be beneficial when strengthened or made adaptive. Updating $\pi_{\text{ref}}$ alongside the policy helps mitigate over-optimization \citep{gorbatovski2024learn, he2026unifying}, and employing a stronger or better-matched reference improves stability and final performance, sometimes surpassing reference-free methods \citep{liu2024understanding, azar2024general, pan2025pre}. This leaves a practical dilemma: removing $\pi_{\text{ref}}$ eliminates metric mismatch and can enhance performance, whereas retaining $\pi_{\text{ref}}$ improves training stability and can also enhance performance.

\begin{figure*}[t]
 \begin{subfigure}[b]{0.31\textwidth}
     \includegraphics[width=\textwidth]{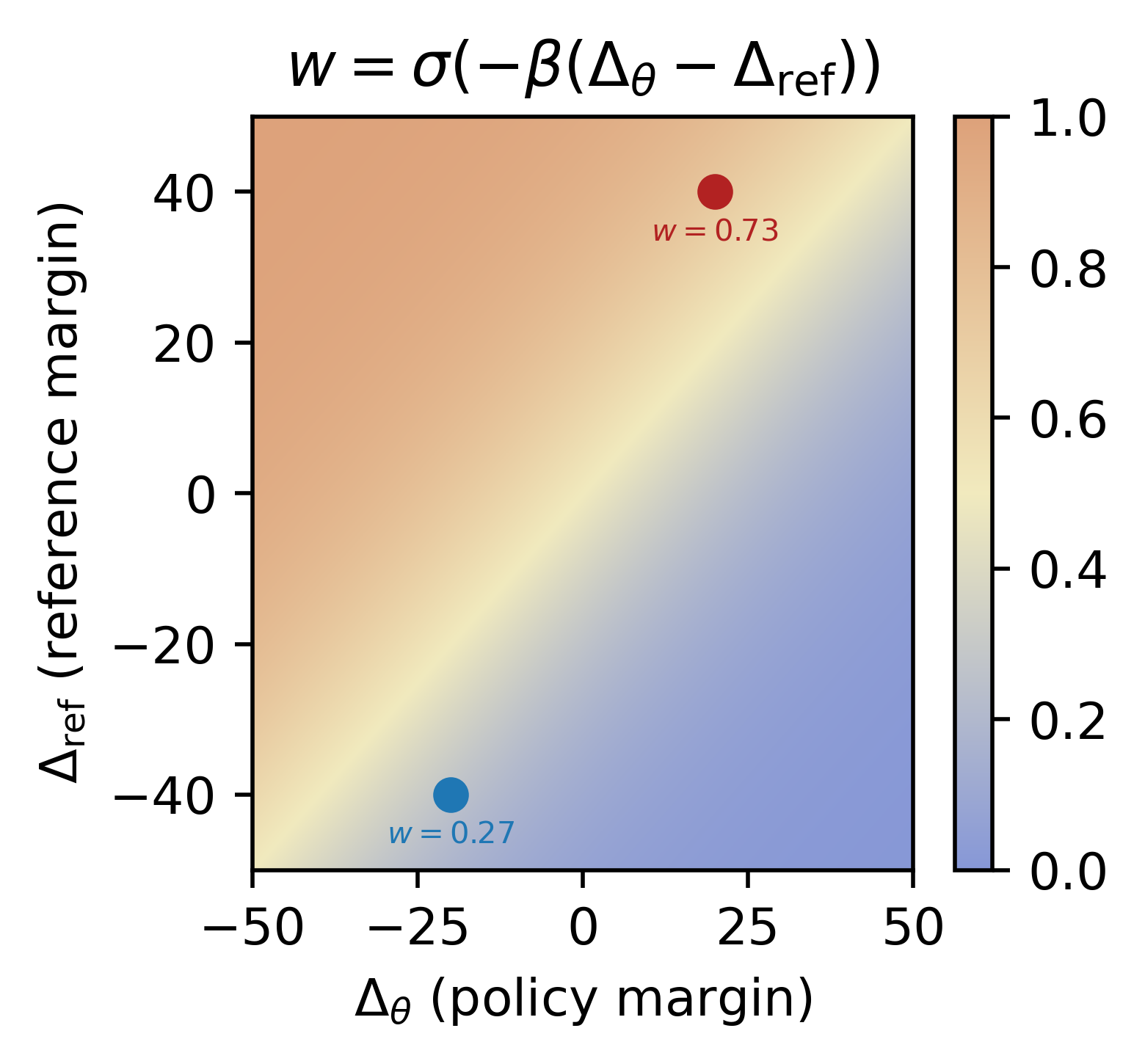}
     \vskip -0.05in
     \caption{DPO}
 \end{subfigure}
 \hfill
 \begin{subfigure}[b]{0.31\textwidth}
     \includegraphics[width=\textwidth]{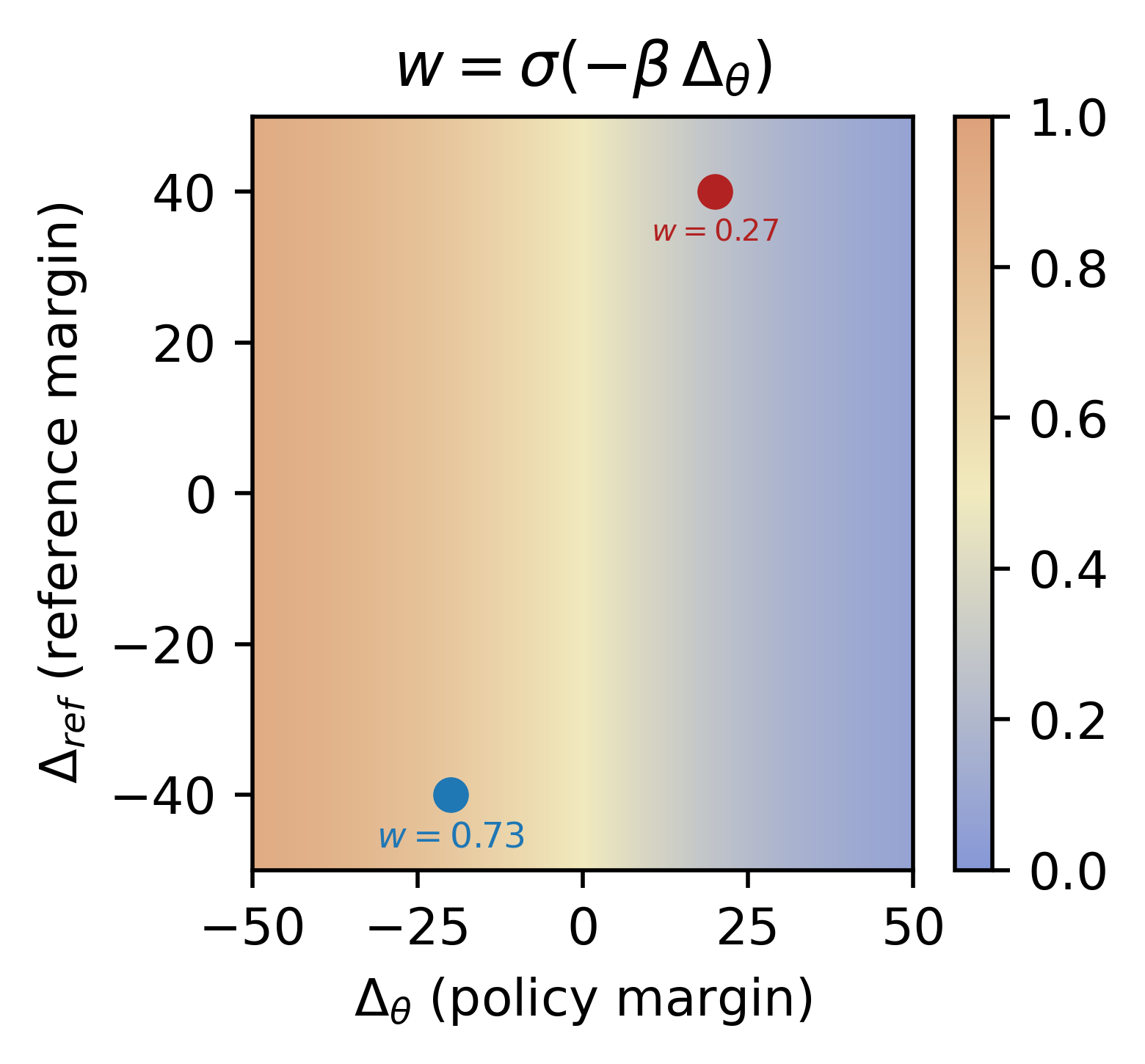}
     \vskip -0.05in
     \caption{Reference-free Method}
 \end{subfigure}
 \hfill
 \begin{subfigure}[b]{0.31\textwidth}
     \includegraphics[width=\textwidth]{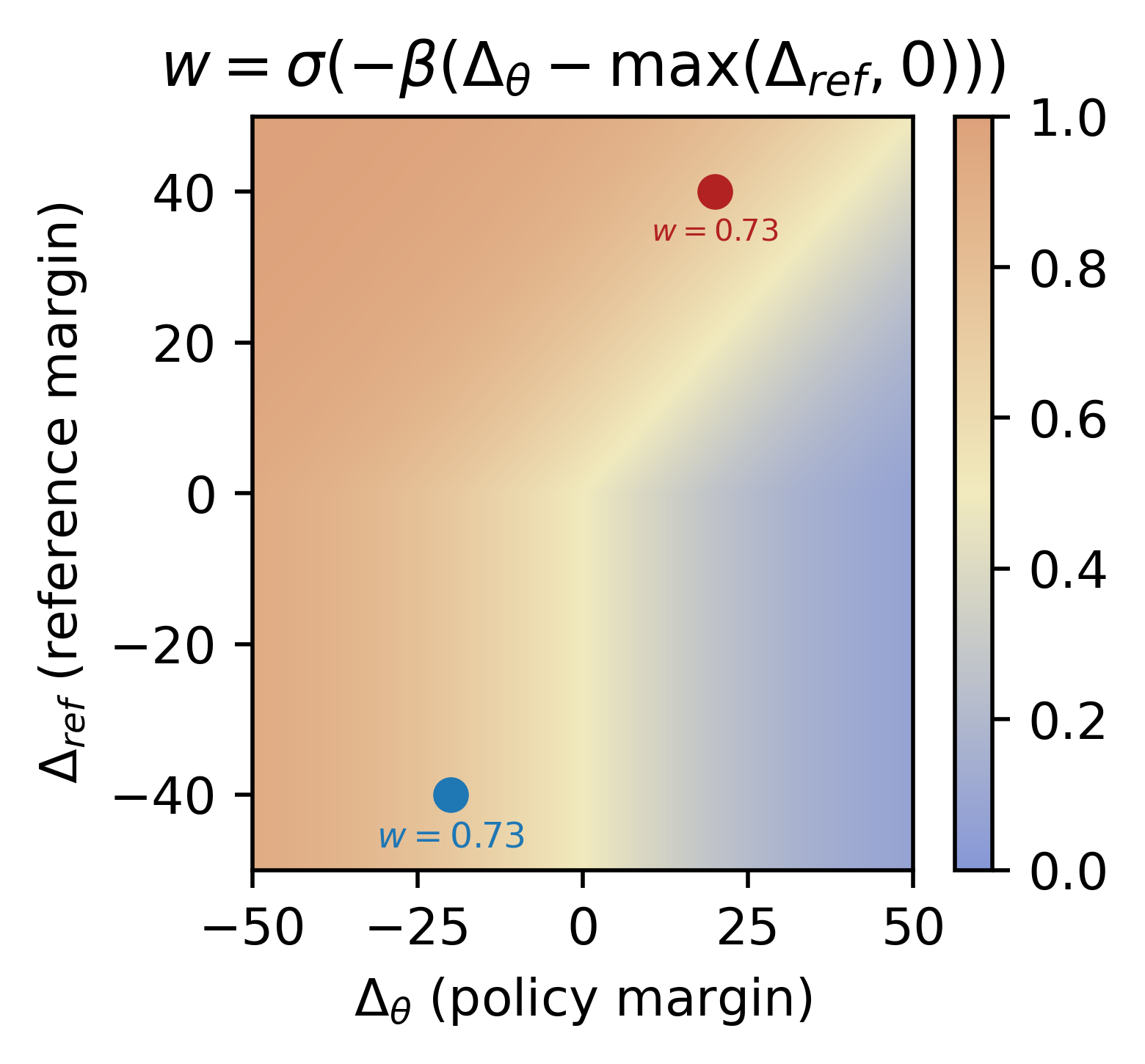}
     \vskip -0.05in
     \caption{\emph{HyPO} (Ours)}
 \end{subfigure}
 \vskip -0.07in
\caption{Per-example gradient weight heatmaps. The weight, plotted over the policy margin $\Delta_\theta$ ($x$-axis) and the reference margin $\Delta_{\mathrm{ref}}$ ($y$-axis), indicates training signal strength. DPO’s reliance on the relative margin leads to \emph{premature satisfaction}: on pessimistic examples (blue dot, $\Delta_{\mathrm{ref}}<0$), the signal is heavily attenuated even though the policy is still wrong in absolute terms. Reference-free methods provide a strong signal but discard the reference entirely. \emph{HyPO} \textcolor{changecolor}{mitigates} this by mirroring the reference-free behavior on pessimistic examples to ensure a strong signal, while reverting to DPO on optimistic examples (red dot) to maintain proximity to the reference policy.}
\label{figure1}
\vskip -0.16in
\end{figure*}

To \textcolor{changecolor}{mitigate} the tension between reference-based and reference-free direct alignment, we propose a unifying principle: the proximal stability offered by the reference model should be applied conditionally. We argue that the reference provides a constructive signal for stable learning only when it aligns with the preference data. When it is pessimistic ($\Delta_{\mathrm{ref}}<0$), its stabilizing pull becomes a misleading bias, making the learning objective unnecessarily difficult and causing the gradient to attenuate even when the absolute ordering ($\Delta_\theta$) is still incorrect. We term this effect \emph{premature satisfaction}. Prior approaches either accept this bias (standard DPO; \citealp{rafailov2023direct}) or discard potentially useful reference signal altogether (reference-free methods; \citealp{yuan2023rrhf, meng2024simpo}).

To realize this principle, we introduce \emph{Hybrid-DPO} (\emph{HyPO}), a minimalist and efficient modification to the DPO objective. \emph{HyPO} behaves identically to DPO when the reference is helpful (optimistic or neutral) but treats a pessimistic reference as neutral ($\Delta_{\mathrm{ref}}=0$). This is achieved with a minor change to the DPO loss: the log-likelihood margin $\Delta_\theta-\Delta_{\mathrm{ref}}$ becomes $\Delta_\theta-\max\{0,\Delta_{\mathrm{ref}}\}$. As illustrated in Figure~\ref{figure1}, standard DPO (a) prematurely attenuates the learning signal on pessimistic examples; typical reference-free method (b) discard the reference signal entirely; our proposed \emph{HyPO} (c) uses the stabilizing reference where it helps and nullifies it where it is detrimental, thereby continuing to push the policy toward the correct absolute ordering.

We conduct a comprehensive empirical evaluation on preference datasets, aligning both base and instruction-tuned variants of two widely-used LLM family \citep{jiang2023clip, llama3modelcard}. Our proposed method, \emph{HyPO}, demonstrates superior performance on general instruction-following benchmarks like AlpacaEval 2.0 \citep{alpaca_eval} and Arena-Hard-v0.1 \citep{arenahard2024}. Notably, these results translate to a 41.2\% average relative improvement over the original DPO. Further analyses and ablations corroborate the robustness of our approach, showing that \emph{HyPO} maintains its advantage when scaling to larger models and transferring to different datasets, all while preserving performance on a wide array of downstream tasks. These findings empirically validate our central thesis: conditionally debiasing the pessimistic reference signal, rather than discarding it entirely, provides a more robust and effective path toward stable, high-performing direct preference alignment.

Our main contributions can be summarized as follows:
\begin{itemize}[leftmargin=0.4cm,topsep=-2pt]
\item [1.]
We identify and formalize \emph{premature satisfaction}: for pessimistic pairs ($\Delta_{\mathrm{ref}}<0$), the DPO weight attenuates once the \emph{relative} margin turns positive even when the \emph{absolute} margin remains wrong ($\Delta_\theta<0$), revealing a tension between correctness and stability.
\item [2.]
We propose \emph{Hybrid-DPO} (\emph{HyPO}), a minimalist modification to the DPO objective that clips the reference margin in the DPO loss, thereby preserving the DPO-style logistic loss and, on pessimistic pairs ($\Delta_{\mathrm{ref}}<0$), reducing to an absolute margin update, with no additional computational cost.
\item [3.]
Across large-scale preference alignment, \emph{HyPO} improves inference-aligned metrics and raises pairwise win rates on general instruction-following 
benchmarks (e.g., AlpacaEval, Arena-Hard).
\end{itemize}

\section{Related Work}
\label{sec:related-work}

We situate \emph{HyPO} within the off-policy branch of RLHF: methods trained on fixed human preference pairs without on-policy rollouts. Accordingly, we do not compare to on-policy RLHF \citep{christiano2017deep, shao2024deepseekmath}, and instead focus on recent direct preference alignment (DPO-style) objectives and extensions, underscoring \emph{HyPO}’s distinct features.

\emph{Manipulating the Reference Signal.}
DPO’s reliance on $\pi_{\mathrm{ref}}$ has inspired a spectrum of solutions. One direction focuses on refining the reference: performance is sensitive to reference choice and strength \citep{liu2024understanding, azar2024general}, leading to Trust-Region DPO (TR-DPO) \citep{gorbatovski2024learn}, which dynamically updates $\pi_{\mathrm{ref}}$ during training to keep it aligned with the policy. RainbowPO \citep{zhao2025rainbowpo} mixes the standard reference with an idealized constant-margin policy. Pre-DPO \citep{pan2025pre} further refines the reference model's role as an implicit sample reweighter. At the other extreme, reference-free methods like RRHF \citep{yuan2023rrhf}, SLiC-HF \citep{zhao2023slic}, CPO \citep{xu2024contrastive}, ORPO \citep{hong2024orpo}, and SimPO \citep{meng2024simpo} eliminate the reference to directly optimize absolute log-likelihoods, addressing the \emph{training–inference mismatch} at the cost of removing a stabilizing prior. The core problem of reference mismatch has also been identified in other modalities; for example, MaPO \citep{she2024mapo} removes the reference to improve multimodal alignment. \emph{HyPO} offers a principled middle ground: it keeps the DPO-style logistic loss, but clips the reference margin, replacing $\Delta_{\mathrm{ref}}$ with $\max\{0,\Delta_{\mathrm{ref}}\}$. This removes pessimistic anchoring while preserving proximality when the reference is non-pessimistic.

\emph{Explicit Reweighting and Robust Optimization.}
A parallel line of work makes the reference-induced sample weighting explicit to improve training dynamics and robustness. Several distributionally robust DPO variants (e.g., rDPO \citep{chowdhury2024provably}, Dr.DPO \citep{wu2024towards}, WDPO \citep{xu2025robust}) reweight examples to minimize the impact of label noise \citep{yuan2023late, yuan2024early, yuan2025instancedependent, yuan2025enhancing} or worst-case distributional shifts. Others use data-driven signals; for instance, $\beta$-DPO \citep{wu2024beta} dynamically adjusts the temperature $\beta$ based on data quality, while WPO \citep{zhou2024wpo} reweights pairs to better simulate on-policy optimization from off-policy data. Some methods operate at a finer granularity, such as TDPO \citep{zeng2024token} and TIS-DPO \citep{liu2024tis}, which apply importance sampling at the token level. While these methods all leverage reweighting, their criteria are typically noise, data quality, or on-policy correction. In contrast, \emph{HyPO} gates the reference contribution by the reference model’s own pessimism: when $\Delta_{\mathrm{ref}}\ge 0$ it behaves exactly like DPO, and when $\Delta_{\mathrm{ref}}<0$ it reduces to an absolute margin update on those pairs.

\emph{Alternative Objective Formulations.}
Beyond manipulating the reference or loss weights, several methods alter the structure of the DPO objective. $f$-DPO generalizes DPO beyond reverse-KL to broader $f$-divergences \citep{wangbeyond}. R-DPO diagnoses and mitigates length biases within the DPO objective \citep{park2024disentangling}, \textcolor{changecolor}{and AlphaDPO adaptively sets a data-dependent reward margin \citep{wualphadpo}.}
 Techniques such as DPO-Shift \citep{yang2025dpo} and BDPO \citep{cho2025rethinking} modify the loss function to bound or rebalance the influence of the rejected response, while FocalPO \citep{liu-etal-2025-focalpo} adds a modulating factor to focus on correctly ranked ones. Anchored methods like APO \citep{d2024anchored} retain the DPO form but add explicit constraints to control the absolute likelihood drift of chosen or rejected responses. 
\emph{HyPO} differs in where it acts: it preserves DPO’s gradient form (same logistic weighting and $\beta$) but modifies the argument of the loss to $\Delta_\theta-\tilde{\Delta}{\mathrm{ref}}$ with $\tilde{\Delta}{\mathrm{ref}}=\max\{0,\Delta_{\mathrm{ref}}\}$, thereby reallocating gradient mass on pessimistic pairs without introducing an extra multiplicative weight.

\section{Mitigating Mismatch in Direct Preference Alignment}

\label{sec3_1}
To fully appreciate the design trade-offs in modern alignment methods, we briefly recall the KL-regularized RLHF principle \citep{christiano2017deep, ziegler2019fine} that underpins them. We begin by considering a distribution of prompts $\mathcal{D}_{p}$ from which prompts $x$ are sampled, and let $\pi_\theta$ be the trainable policy initialized from a supervised policy $\pi_{\text{SFT}}$. In KL-regularized policy optimization (e.g., PPO; \citealp{schulman2017proximal}), a canonical KL-regularized objective used in practice is:
\begin{equation}
\label{eq:kl-rlhf}
\max_{\pi_\theta}\;
\mathbb{E}_{x\sim \mathcal{D}_{p},\, y\sim \pi_\theta(\cdot\mid x)}
\big[r_\phi(x,y)\big]
\;-\;
\tau\,\mathbb{E}_{x\sim\mathcal{D}_{p}}\!\big[D_{\mathrm{KL}}\big(\pi_\theta(\cdot\mid x)\;\|\;\pi_{\text{ref}}(\cdot\mid x)\big)\big],
\end{equation}
where $r_\phi$ is a reward model trained from human preferences, $\pi_{\text{ref}}$ is a fixed reference (typically $\pi_{\text{SFT}}$), and $\tau>0$ controls the regularization strength.
For each $x$, $\pi_\theta^\star$ denotes the optimal policy, the per-context optimum of \eqref{eq:kl-rlhf} has the Gibbs form \citep{peters2010relative, azar2024general}:
\begin{equation}
\label{eq:gibbs-kl}
\pi_\theta^\star(y\mid x)=
\frac{\pi_{\text{ref}}(y\mid x)\,\exp\!\big(r_\phi(x,y)/\tau\big)}
{\sum_{y'} \pi_{\text{ref}}(y'\mid x)\,\exp\!\big(r_\phi(x,y')/\tau\big)},
\end{equation}
which exposes the role of the KL term: it anchors $\pi_\theta$ near $\pi_{\text{ref}}$, making updates stable and discouraging distributional drift that could exploit reward misspecification \citep{pmlr-v37-schulman15,schulman2017proximal}.

\textbf{Direct Preference Optimization} (DPO; \citealp{rafailov2023direct}) circumvents explicit reward modeling and reinforcement learning by reformulating the objective for an offline setting, directly optimizing pairwise preferences $(x,y^+,y^-)$ where $y^+\succ y^-$ under a Bradley--Terry likelihood \citep{bradley1952rank}. We define the policy and reference margins as:
\vspace{-0.3em}\begin{equation}
\label{eq:margins}
\begin{aligned}
\Delta_\theta(x,y^+,y^-) &= \log \pi_\theta(y^+\mid x)-\log \pi_\theta(y^-\mid x),\\
\Delta_{\mathrm{ref}}(x,y^+,y^-) &= \log \pi_{\mathrm{ref}}(y^+\mid x)-\log \pi_{\mathrm{ref}}(y^-\mid x).
\end{aligned}
\end{equation}\vspace{-0.3em}
DPO minimizes the logistic loss:
\begin{equation}
\label{eq:dpo-loss}
\mathcal{L}_{{\text{DPO}}}
=\mathbb{E}_{(x,y^+,y^-)}\!\left[-\log \sigma\!\Big(\beta\big(\Delta_\theta-\Delta_{\mathrm{ref}}\big)\Big)\right]
=\mathbb{E}_{(x,y^+,y^-)}\!\left[\log\!\Big(1+\exp\!\big(-\beta(\Delta_\theta-\Delta_{\mathrm{ref}})\big)\Big)\right],
\end{equation}
where $\beta=1/\tau$. Under the Bradley--Terry model, the first-order optimality condition shares the Gibbs structure of KL-regularized policy optimization \citep{rafailov2023direct} and satisfies:
\begin{equation}
\label{eq:gibbs-consistency}
\log \pi_\theta(y\mid x)-\log \pi_{\mathrm{ref}}(y\mid x)
= \beta\, r^\star(x,y) - \log Z(x).
\end{equation}
This follows by taking logarithms of the KL-regularized optimum in Eq. \ref{eq:gibbs-kl}:
\begin{equation}
\label{eq:dpokl}
\pi^\star(y\mid x)\ \propto\ \pi_{\mathrm{ref}}(y\mid x)\,\exp\!\big(r^\star(x,y)/\beta\big),
\end{equation}
where $r^\star$ is a preference-consistent reward and $Z(x)$ the partition function. This makes explicit how DPO preserves proximity to $\pi_{\mathrm{ref}}$ without explicit RL.
Differentiating Eq.~\ref{eq:dpo-loss} with respect to the policy margin $\Delta_\theta$ yields the per-example gradient-weight form:
\begin{equation}
\begin{aligned}
\label{eq:dpo-grad}
\nabla_\theta \ell_{\mathrm{DPO}}
=-\beta\,\underbrace{\sigma\!\big(-\beta(\Delta_\theta-\Delta_{\mathrm{ref}})\big)}_{w_{\mathrm{DPO}}\in(0,1)}\,\nabla_\theta \Delta_\theta.
\end{aligned}
\end{equation}\vspace{-1.2em}

Although the introduction of the reference policy anchors the policy \(\pi_\theta\) near \(\pi_{\mathrm{ref}}\) via the relative margin \((\Delta_\theta-\Delta_{\mathrm{ref}})\) serving as the gradient weight \citep{pan2025pre} in Eq.~\ref{eq:dpo-grad}, thereby providing stability and suppressing distributional drift, it simultaneously introduces a \emph{training--inference metric mismatch} \citep{chen2024preference, meng2024simpo}. We refer to this side effect as \emph{premature satisfaction}.

\textbf{Premature Satisfaction.}
Let the scaled relative margin be:
\begin{equation}
z=\beta\big(\Delta_\theta-\Delta_{\mathrm{ref}}\big),\qquad
w_{\mathrm{DPO}}=\sigma(-z)=\frac{1}{1+e^{z}},\quad
z\ge 0\ \Rightarrow\ w_{\mathrm{DPO}}\le e^{-z}.
\end{equation}
Hence, from Eq.~\ref{eq:dpo-grad} we obtain:
$\bigl\|\nabla_\theta \ell_{\mathrm{DPO}}\bigr\|
=\beta\,\sigma(-z)\,\bigl\|\nabla_\theta \Delta_\theta\bigr\|
\ \le\ \beta\,e^{-z}\,\bigl\|\nabla_\theta \Delta_\theta\bigr\|.$

When the reference is pessimistic on a pair (\(\Delta_{\mathrm{ref}}<0\)), even if the current policy is still wrong in absolute terms (\(\Delta_\theta<0\)), as soon as the relative margin becomes positive (\(\Delta_\theta-\Delta_{\mathrm{ref}}>0\)), the gradient shrinks exponentially, even though the policy remains wrong in absolute terms (\(\Delta_\theta<0\)).

For example, take \(\beta=1\), \(\Delta_{\mathrm{ref}}=-3\), and \(\Delta_\theta=-1\). Then \(\Delta_\theta-\Delta_{\mathrm{ref}}=2\), we have 
$w_{\mathrm{DPO}}=\sigma(-2)\approx 0.119,
\bigl\|\nabla_\theta \ell_{\mathrm{DPO}}\bigr\|\approx 0.119\,\bigl\|\nabla_\theta \Delta_\theta\bigr\|.
$
As illustrated by the lower-left region of the heatmap in Figure~\ref{figure1}(a) (i.e., \(\Delta_\theta-\Delta_{\mathrm{ref}}>0\)), the weight \(w_{\mathrm{DPO}}\) decays rapidly; when \(\Delta_{\mathrm{ref}}<0\), this low-weight band still covers many points with \(\Delta_\theta<0\).
\emph{Premature satisfaction} reveals the tension between stability (proximality) and mismatch: while DPO suppresses drift, it may also systematically attenuate the training signal of examples that disagree with \(\pi_{\mathrm{ref}}\) yet remain substantially wrong.

\textbf{Reference-free Alignment.} A natural reaction is to remove the reference to obtain an unbiased signal based on the absolute correctness of $\Delta_\theta$, aligned with inference-time likelihoods. Reference-free objectives (e.g., SimPO; \citealp{meng2024simpo}) optimize the absolute margin:
\begin{equation}
\label{eq:abs-loss}
\mathcal{L}_{\text{abs}}
=\mathbb{E}_{(x,y^+,y^-)}
\Big[\;\ell_{\text{abs}}(\Delta_\theta)\;\Big],
\qquad
\ell_{\text{abs}}(\Delta_\theta)=\log\!\big(1+\exp(-\beta\,\Delta_\theta)\big),
\end{equation}
with gradient:
\begin{equation}
\label{eq:abs-grad}
\frac{\partial \ell_{\text{abs}}}{\partial \Delta_\theta}
=-\beta\,\sigma(-\beta\Delta_\theta),
\qquad
\nabla_\theta \ell_{\text{abs}}
=-\beta\,\sigma(-\beta\Delta_\theta)\,\nabla_\theta \Delta_\theta.
\end{equation}
Compared to Eq.~\ref{eq:dpo-grad}, attenuation now depends only on the policy’s own performance: pairs with $\Delta_\theta<0$ inevitably receive large gradients until $\Delta_\theta$ crosses zero, directly aligning the training signal with the inference-time objective (increasing $\log\pi_\theta(y^+\mid x)$ and decreasing $\log\pi_\theta(y^-\mid x)$). However, Eq.~\ref{eq:abs-loss} discards the very mechanism that implemented proximality in Eq.~\ref{eq:dpo-loss} and Eq.~\ref{eq:gibbs-consistency}. Although recent work suggests that RLHF \citep{chu2025sft, mukherjee2025reinforcement} does not typically induce substantial parameter drift, removing this proximality nonetheless markedly reduces the robustness \citep{pan2025pre, liu2024understanding} of reference-free alignment.

\textbf{Direct Preference Optimization with Better Reference.}
Another approach argues that we should improve reference model to enhance DPO framework. This idea motivates using a stronger reference policy within the DPO framework \citep{liu2024understanding, azar2024general, pan2025pre}. The premise is that a more accurate reference margin, $\Delta_{\mathrm{ref}}$, makes the relative margin $(\Delta_\theta - \Delta_{\mathrm{ref}})$ a better learning signal, which should improve optimization.

\begin{wrapfigure}{r}{0.32\textwidth}
\vskip -0.17in
  \centering
  \includegraphics[width=0.32\textwidth]{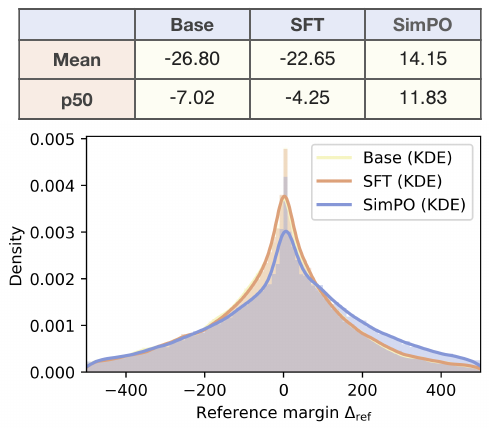}
  \vskip -0.05in
  \caption{Distribution of the reference margin ($\Delta_{\mathrm{ref}}$) across different reference models. The table reports the mean and median (p50) of $\Delta_{\mathrm{ref}}$ for each model.}
\vskip -0.05in
\label{figure2}
\end{wrapfigure}

While often effective, these methods only modify the reference policy $\pi_{\mathrm{ref}}$ and leave the core DPO loss unchanged. Consequently, the phenomenon we term \emph{premature satisfaction} persists whenever the reference is pessimistic on a pair (i.e., $\Delta_{\mathrm{ref}}<0$). To quantify this, we compute the sequence-level likelihood margin in Eq.~\eqref{eq:margins} under three Llama-3-8B–based references (Base, SFT, and a SimPO-aligned model) on the UltraFeedback training split \citep{cui2023ultrafeedback}, and report (i) the density of $\Delta_{\mathrm{ref}}$ and (ii) summary statistics (full settings and results in Appendix~\ref{appendix_settings}). Stronger references are indeed pessimistic less often and shift the distribution of $\Delta_{\mathrm{ref}}$ to the right; yet a substantial fraction, approximately $45\%$, of pairs remains in the pessimistic region even for the SimPO-aligned reference, which is designed to mitigate such mismatch. This imposes a practical ceiling on the “better reference” strategy: it improves stability and overall performance, but the \emph{training–inference mismatch} persists wherever $\Delta_{\mathrm{ref}}<0$.

\subsection{Hybrid-DPO}

\label{sec:hypo}

In this section, we introduce \emph{Hybrid-DPO} (\emph{HyPO}), an approach designed to mitigate the \emph{premature satisfaction} problem while preserving the stability of original DPO \citep{rafailov2023direct}. The stability benefit of DPO, inherited from KL-regularized RLHF (Eq.~\ref{eq:dpokl}), relies on the assumption that the reference policy $\pi_{\mathrm{ref}}$ provides a constructive learning signal. We argue that this KL-induced proximality should be \emph{conditional}: When the reference is optimistic or neutral ($\Delta_{\mathrm{ref}} \ge 0$), it provides a valid anchor for stable, relative improvement, as the DPO objective effectively preserves proximity to the reference (Eq.~\ref{eq:gibbs-consistency}). However, when the reference is pessimistic ($\Delta_{\mathrm{ref}} < 0$), it acts as a misleading prior that causes \emph{premature satisfaction}; as analyzed in our discussion of the DPO gradient, the weight $w_{\mathrm{DPO}}$ in Eq.~\ref{eq:dpo-grad} attenuates even when the absolute margin $\Delta_\theta$ remains negative. \emph{HyPO} realizes this principle by clipping the reference margin: it preserves DPO's relative objective (relative margin, $(\Delta_\theta - \Delta_{\mathrm{ref}})$) for reliable pairs but nullifies the reference's influence for pessimistic ones, compelling the model to optimize the absolute objective (absolute margin, $\Delta_\theta$) instead.

\textbf{Objective.}
Formally, for a given threshold $\gamma \ge 0$ (with $\gamma=0$ by default), we modify the DPO objective by replacing the reference margin $\Delta_{\mathrm{ref}}$ with its clipped, non-negative counterpart:
\begin{equation}
\label{eq:hypo-margin-hard}
\widetilde{\Delta}_{\mathrm{ref}} = \max\{\Delta_{\mathrm{ref}}, \gamma\}, \quad \text{and} \quad
\mathcal{L}_{\mathrm{HyPO}} = \mathbb{E}\left[\log\left(1 + \exp\big(-\beta(\Delta_\theta - \widetilde{\Delta}_{\mathrm{ref}})\big)\right)\right].
\end{equation}
The gradient of the per-example loss retains the DPO structure, weighted by the modified margin:
\begin{equation}
\label{eq:hypo-grad}
\nabla_\theta \mathcal{L}_{\mathrm{HyPO}} = -\beta \cdot \underbrace{\sigma\big(-\beta(\Delta_\theta - \widetilde{\Delta}_{\mathrm{ref}})\big)}_{w_{\mathrm{HyPO}}} \cdot \nabla_\theta \Delta_\theta.
\end{equation}
By construction, \emph{HyPO} coincides with DPO on \emph{non-pessimistic} samples ($\Delta_{\mathrm{ref}}\!\ge\!\gamma$), i.e., its per-example weight in Eq.~\ref{eq:hypo-grad} equals that of DPO in Eq.~\ref{eq:dpo-grad} ($w_{\mathrm{HyPO}}\!=\!w_{\mathrm{DPO}}$), thereby preserving proximal stability where the reference is reliable. On pessimistic samples ($\Delta_{\mathrm{ref}}\!<\!\gamma$), clipping nullifies the misleading anchor and restores an absolute–margin–driven update, pointwise dominating the reference-free weight $w_{\mathrm{abs}}:=\sigma(-\beta\Delta_\theta)$ induced by Eq.~\ref{eq:abs-grad} ($w_{\mathrm{HyPO}}\!\ge\!w_{\mathrm{abs}}$), thus preventing premature attenuation. An intuitive visualization is shown in Figure~\ref{figure1}(c).

For a globally smooth objective, we can replace the hard maximum with a softplus function \citep{dugas2000incorporating}. The smoothed reference margin is defined as:
\begin{equation}
\label{eq:hypo-softplus}
\widetilde{\Delta}_{\mathrm{ref}}
=\gamma+\frac{1}{\alpha}\log\!\big(1+\exp(\alpha(\Delta_{\mathrm{ref}}-\gamma))\big),\quad \alpha>0,
\end{equation}
where $\alpha$ controls the smoothness ($\alpha \to \infty$ recovers Eq.~\ref{eq:hypo-margin-hard}). This simple, smoothed formulation allows \emph{HyPO} to be implemented as a plug-in modification to the standard DPO loss.

\begin{figure*}[h]
\vskip -0.0in
 \begin{subfigure}[b]{0.26\textwidth}
     \includegraphics[width=\textwidth]{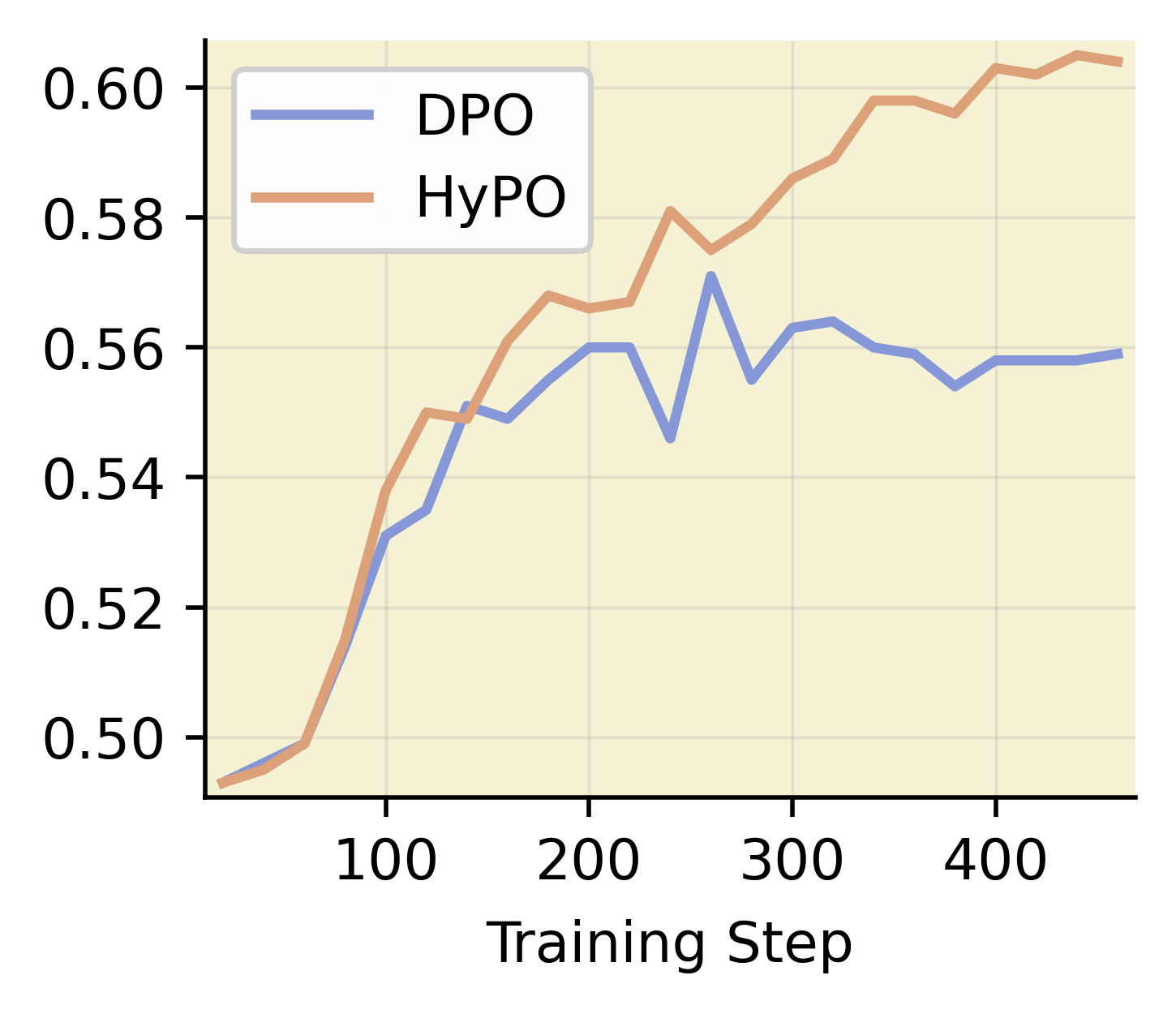}
     \vskip -0.1in
     \caption{Agreement Rate}
 \end{subfigure}
 \hfill
 \begin{subfigure}[b]{0.26\textwidth}
     \includegraphics[width=\textwidth]{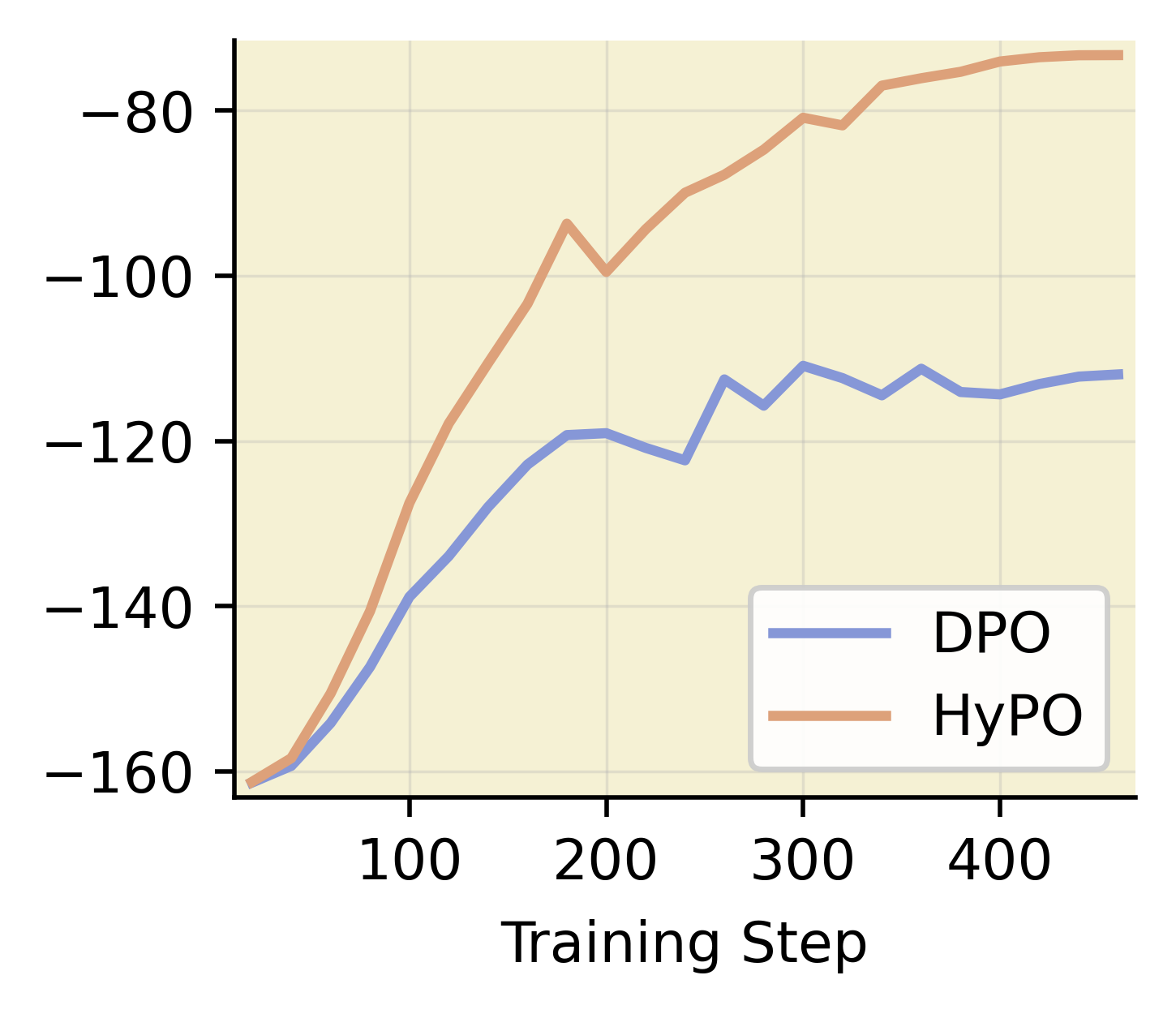}
     \vskip -0.1in
     \caption{Absolute Margin}
 \end{subfigure}
 \hfill
 \begin{subfigure}[b]{0.466\textwidth}
     \includegraphics[width=\textwidth]{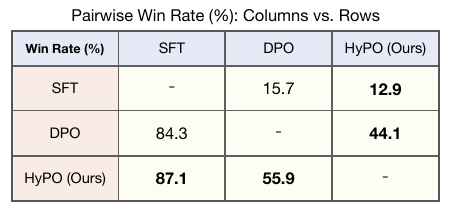}
     \vskip -0.05in
     \caption{Win-Rate Matrix}
 \end{subfigure}
\caption{\emph{HyPO} improves inference-aligned evaluation metrics and pairwise win rates.
(a) \emph{Absolute agreement rate} over training (higher is better). 
(b) \emph{Absolute margin on the pessimistic subset} ($\Delta_{\mathrm{ref}}<0$). 
(c) \emph{Pairwise win-rate}. Each cell is the win rate (\%) of the row model against the column model on AlpacaEval 2.0 \citep{alpaca_eval}.
All results use the same SFT checkpoint of Llama-3-8B-Base~\citep{llama3modelcard} trained on UltraFeedback~\citep{cui2023ultrafeedback} with either DPO or our \emph{HyPO}; the training/evaluation pipeline and optimization hyperparameters are identical and set to the DPO configuration from Zephyr~\citep{tunstall2023zephyr}. 
See Section \ref{settings} for more settings.}
\label{figure3}
\vskip -0.0in
\end{figure*}

\textbf{Effectiveness of \emph{HyPO}.} To quantify how effectively the \emph{HyPO} objective mitigates the \emph{training–inference mismatch} compared to DPO \citep{rafailov2023direct}, we track two key metrics on the evaluation set that are closely aligned with inference-time performance. The first, the \emph{Absolute Agreement Rate}, serves as a global measure of overall performance. Defined as $\operatorname{Agree}@t := \Pr[\Delta_\theta^{(t)} > 0]$ (with $\Delta_\theta$ the absolute policy margin as in Eq.~\ref{eq:margins}), it measures the fraction of evaluation pairs whose absolute likelihood ordering matches the desired preference, directly mirroring the inference-time decision rule.
Further, we use a more targeted \emph{Absolute Margin on the Pessimistic Subset}, which tracks the average absolute margin, $\mathbb{E}[\Delta_\theta \mid \Delta_{\mathrm{ref}} < 0]$, on the subset where the (fixed) reference policy is pessimistic. This metric isolates the challenging cases that trigger \emph{premature satisfaction} in DPO and precisely measures \emph{HyPO}’s ability to improve learning on them. Further details on metric definitions are provided in Appendix~\ref{appednix_metric}.
To directly compare final model outputs, we report a pairwise win rate matrix using AlpacaEval 2.0 \citep{alpaca_eval}. For models $i$ (row) and $j$ (column), the entry $W_{i,j}:=100\times \Pr[\text{model }i \succ \text{model }j]$
is the percentage of prompts on which the row model’s response is preferred to the column model’s, under the same judge and prompts; diagonals are undefined. Configuration and evaluation protocol are detailed in Section \ref{settings}.

As demonstrated in Figure~\ref{figure3}(a) and (b), \emph{HyPO} achieves a significantly faster and higher increase in both the global agreement rate and the pessimistic subset margin, confirming its effectiveness in mitigating the \emph{training–inference mismatch}. \textcolor{changecolor}{In this run, the final agreement rate increases from $55.9\%$ for DPO to $60.4\%$ for \emph{HyPO}.}
As shown in Figure~\ref{figure3}(c), \emph{HyPO} outperforms \emph{DPO} head-to-head ($W_{\text{HyPO},\text{DPO}}=55.9\%$ vs.\ $W_{\text{DPO},\text{HyPO}}=44.1\%$). Moreover, against the common baseline \emph{SFT}, both alignment methods win, with \emph{HyPO} achieving the larger advantage.

A key advantage of \emph{HyPO} is that it keeps a DPO-like form and is therefore compatible with enhancements for the DPO. In our practical implementation, we leverage this compatibility in two ways. First, we adopt a better reference \citep{liu2024understanding, pan2025pre} by using a pre-aligned, same-family model instead of the SFT default. Second, we apply a uniform home advantage margin $h$ to impose stronger learning pressure, leading to the final objective: $\Delta_\theta \;-\; \max\{\Delta_{\mathrm{ref}},\,\gamma\} \;-\; h.$
We do not use length normalization \citep{park2024disentangling} due to its hyperparameter sensitivity.

\section{Experiments}

We conduct a series of experiments to validate the efficacy of \emph{HyPO}. We first establish its outperformance against a set of offline preference alignment baselines on standard benchmarks. Subsequently, we perform targeted analyses and ablations to probe the robustness and scalability of \emph{HyPO}.

\subsection{Experimental Setup and Results}
\label{settings}
To ensure comprehensive and fair comparison of our proposed \emph{HyPO} against various direct preference alignment baselines, we closely follow the established experimental settings from existing works~\citep{meng2024simpo, rashidinejadsail, gorbatovski2024learn}. More detailed
experimental settings can be found in Appendix~\ref{appendix_settings}.

\textbf{Models and datasets settings.}
Our main experiments evaluate two open-source model families, Mistral-7B \citep{jiang2023clip} and Llama3-8B \citep{llama3modelcard}, using the UltraFeedback dataset ($\approx 61$k training samples; \citealp{cui2023ultrafeedback} for preference alignment. We conduct these experiments in two distinct settings.
In the \emph{Base setting}, we first perform supervised fine-tuning (SFT) on the raw pretrained models (e.g., Mistral-7B-v0.1, Meta-Llama-3-8B) using the UltraChat-200k dataset ($\approx 200$k training samples; \citealp{ding2023enhancing}). The resulting SFT policy then serves as the initial policy for preference alignment on UltraFeedback. In the \emph{Instruct setting}, we bypass the SFT step and directly apply our preference alignment methods to the official instruction-tuned models (e.g., Mistral-7B-Instruct-v0.2, Meta-Llama-3-8B-Instruct), again using the UltraFeedback dataset.

\textbf{Training settings.} Our training methodology is grounded in established practices from recent alignment literature to ensure robustness and fair comparison \citep{tunstall2023zephyr, Tunstall_The_Alignment_Handbook}. For the initial SFT stage, we use a peak learning rate of $2 \times 10^{-5}$. For each reproduced direct preference alignment method, we conduct a small grid search over hyperparameters: the peak learning rate is selected from $\{5 \times 10^{-7}, 1 \times 10^{-6}\}$, while the DPO temperature $\beta$ is chosen from $\{0.01, 0.1\}$ for standard objectives or from $\{2, 10\}$ for methods employing length normalization. For our proposed \emph{HyPO}, we use a checkpoint of the same architecture pre-aligned with SimPO. The home advantage margin was set to $h=10$, while the threshold was kept at its default of $\gamma=0$. Detailed settings are listed in Appendix~\ref{appendix_settings}.
All models are trained for a single epoch using the AdamW optimizer \citep{loshchilov2017decoupled}, and we employ a cosine learning rate schedule with a warm-up phase over the first 10\% of training steps and a global batch size of 128.

\textbf{Evaluation benchmarks.} We primarily assess model performance on two popular instruction-following benchmarks: AlpacaEval 2.0 \citep{alpaca_eval, dubois2024length} and Arena-Hard-v0.1 \citep{li2024crowdsourced, arenahard2024}. Consistent with prior work \citep{rashidinejadsail, gorbatovski2024learn}, we use GPT-4-Preview-1106 as the automated judge to compute the primary metric: the win rate, which is the percentage of times a model's response is preferred over a baseline's. The specific baseline model varies by benchmark: AlpacaEval reports win rates against GPT-4-Preview-1106 itself, whereas Arena-Hard uses GPT-4-0314. Our evaluation protocols follow established practices from \citet{tunstall2023zephyr, meng2024simpo, Tunstall_The_Alignment_Handbook}, with the minor modification of using a more recent vLLM release \citep{kwon2023efficient} for generation.

\textbf{Direct preference alignment baselines.} 
We compare our method against a suite of direct preference alignment baselines. Our comparison starts with two classic off-policy methods: SLiC-HF \citep{zhao2023slic} and DPO \citep{rafailov2023direct}, the latter of which serves as our primary baseline. CPO \citep{xu2024contrastive} and KTO \citep{ethayarajh2024kto} are DPO variants that reshape the loss from, respectively, a contrastive perspective and an asymmetric utility. SimPO \citep{meng2024simpo} is a high-performing reference-free representative that operates on the absolute likelihood margin. FocalPO \citep{liu-etal-2025-focalpo}, TR-DPO \citep{gorbatovski2024learn}, and RainbowPO \citep{zhao2025rainbowpo} are recent DPO improvements: focal reweighting, trust-region constraints on the relative margin for stability, and a unified practical recipe combining length normalization and policy mixing.

\begin{table*}[!h]
\centering
\renewcommand{\arraystretch}{1.10}
\small 
\caption{Main results reported from AlpacaEval 2.0 \citep{alpaca_eval, dubois2024length} and Arena-Hard-v0.1 \citep{li2024crowdsourced, arenahard2024}. LC and WR denote \emph{length-controlled win rate} and \emph{raw win rate}, respectively. The best results are highlighted in bold, and the second-best are underlined.}
\vspace{-0.8em}
\resizebox{\textwidth}{!}{
\begin{tabular}{lccccccc}
\toprule
\multirow{3}{*}{\textbf{Method}} & \multicolumn{3}{c}{\textbf{Mistral-Base (7B)}} & \multicolumn{3}{c}{\textbf{Mistral-Instruct (7B)}} \\ 
\cmidrule(lr){2-4}\cmidrule(lr){5-7}
& \multicolumn{2}{c}{\textbf{AlpacaEval}} & \multicolumn{1}{c}{\textbf{Arena-Hard}}  & \multicolumn{2}{c}{\textbf{AlpacaEval}} & \multicolumn{1}{c}{\textbf{Arena-Hard}}  \\
\cmidrule(lr){2-3}\cmidrule(lr){4-4} \cmidrule(lr){5-6}\cmidrule(lr){7-7}
& {\scriptsize \bf LC (\%)} & {\scriptsize \bf WR (\%)} &  {\scriptsize \bf WR (\%)} & {\scriptsize \bf LC (\%)}  & {\scriptsize \bf WR (\%)} & {\scriptsize \bf WR (\%)} \\
\midrule
SLiC-HF~\citep{zhao2023slic} &  11.6 & 9.1  & 5.4 & 32.4 & 31.2 & 16.9  \\
DPO~\citep{rafailov2023direct} &  22.6 & 18.5  & 7.9 & 35.1 & 31.4 & 15.4  \\
CPO~\citep{xu2024contrastive} &  13.1 & 11.6  & 6.4 & 34.9 & 39.9 & \underline{21.0}  \\
KTO~\citep{ethayarajh2024kto} &  12.9 & 9.3  & 6.6 & 35.0 & 31.3 & 17.5  \\
SimPO~\citep{meng2024simpo} &  27.3 & 25.4  & \underline{11.2} & \underline{38.4} & \underline{40.0} & 20.5  \\
FocalPO~\citep{liu-etal-2025-focalpo} &  25.8 & 19.7  & 8.2 & 35.9 & 35.0 & 18.7 \\
TR-DPO~\citep{gorbatovski2024learn} &  24.9 & 21.4  & 9.5 & 36.5 & 33.7 & 18.2 \\
RainbowPO~\citep{zhao2025rainbowpo} &  \underline{28.4} & \underline{26.7}  & 9.2 & 35.7 & 33.9 & 18.2  \\
\midrule
\emph{HyPO}~(Ours) &  \textbf{32.8} & \textbf{29.6}  & \textbf{13.9} & \textbf{38.9} & \textbf{47.9} & \textbf{25.2}  \\
\midrule[.7pt]
\multirow{3}{*}{\textbf{Method}} & \multicolumn{3}{c}{\textbf{Llama-3-Base (8B)}} & \multicolumn{3}{c}{\textbf{Llama-3-Instruct (8B)}} \\ 
\cmidrule(lr){2-4}\cmidrule(lr){5-7}
& \multicolumn{2}{c}{\textbf{AlpacaEval}} & \multicolumn{1}{c}{\textbf{Arena-Hard}}  & \multicolumn{2}{c}{\textbf{AlpacaEval}} & \multicolumn{1}{c}{\textbf{Arena-Hard}}  \\
\cmidrule(lr){2-3}\cmidrule(lr){4-4} \cmidrule(lr){5-6}\cmidrule(lr){7-7}
& {\scriptsize \bf LC (\%)} & {\scriptsize \bf WR (\%)} &  {\scriptsize \bf WR (\%)} & {\scriptsize \bf LC (\%)}  & {\scriptsize \bf WR (\%)} & {\scriptsize \bf WR (\%)} \\
\midrule
SLiC-HF~\citep{zhao2023slic} &  19.8 & 15.9  & 14.3 & 36.7 & 36.8 & 25.1 \\
DPO~\citep{rafailov2023direct} &  24.3 & 21.9  & 23.0 & 40.9  & 41.3 & 31.5  \\
CPO~\citep{xu2024contrastive} &  22.3 & 24.6  & 12.2 & 38.1 & 40.4 & 30.0  \\
KTO~\citep{ethayarajh2024kto} &  23.6 & 20.3  & 18.4 & 40.5 & 39.0 & 30.5  \\
SimPO~\citep{meng2024simpo} &  30.7 & 26.2  & 30.1 & 46.0 & 43.1 & 32.1  \\
FocalPO~\citep{liu-etal-2025-focalpo} &  27.2 & 25.4  & 27.9 & 45.1 & \underline{43.6} & 30.2 \\
TR-DPO~\citep{gorbatovski2024learn} &  \underline{31.8} & \underline{30.2}  & \underline{31.0} & \underline{46.7} & 42.7 & \underline{32.5}  \\
RainbowPO~\citep{zhao2025rainbowpo} &  30.3 & 27.1  & 28.6 & \underline{46.7} & 43.5 & 31.3  \\
\midrule
\emph{HyPO}~(Ours) &  \textbf{34.7} & \textbf{33.6}  & \textbf{33.5} & \textbf{49.5} & \textbf{46.2} & \textbf{35.2}  \\
\bottomrule
\end{tabular}
}
\label{tab:main_res}
\vspace{-0.2em}
\end{table*}

\textbf{Main Results on Benchmarks.} We present the main results of our comparative analysis in Table~\ref{tab:main_res}. Across all four experimental settings, including two model families (Mistral-7B and Llama-3-8B) and two initial states (Base and Instruct), our proposed method, \emph{HyPO}, consistently and significantly outperforms all baselines on both AlpacaEval and Arena-Hard benchmarks. The results show that \emph{HyPO} achieves an average relative improvement of 41.2\% over the original DPO. Furthermore, when compared to SimPO, a strong reference-free competitor, \emph{HyPO} delivers an average relative improvement of 15.1\%. Taken together, these results highlight the practical value of our methodology. By mitigating the reference mismatch while preserving the core DPO framework, our method translates a targeted theoretical improvement into superior performance on benchmarks.

\subsection{Further Analysis and Ablations}
To further validate the effectiveness and robustness of \emph{HyPO}, we evaluate downstream transfer, run ablations (including sensitivity to $\gamma$), measure training runtime, and study scaling. For comparability, we benchmark against one reference-based baseline (DPO; \citealp{rafailov2023direct}) and one reference-free baseline (SimPO; \citealp{meng2024simpo}) under largely the same settings as Section~\ref{settings}.

\textbf{Downstream tasks.}
One of the challenges in alignment is that aligning a model can reduce a model's general capabilities. This degradation is often attributed to the over-optimization \citep{gao2023scaling, rafailov2024scaling} of the alignment objective. To assess whether our approach preserves broad utility, we evaluated the aligned models on a comprehensive suite of downstream tasks, using LM Evaluation Harness \citep{biderman2024lessons}.
Our evaluation covers four key domains. We test knowledge and language understanding using MMLU \citep{hendrycks2020measuring} and ARC-Challenge \citep{clark2018think}. For commonsense reasoning, we employ HellaSwag \citep{zellers2019hellaswag} and Winogrande \citep{sakaguchi2021winogrande}. Factuality is assessed with TruthfulQA \citep{lin2021truthfulqa}, and mathematical reasoning is measured by GSM8K \citep{cobbe2021training}.
The results, presented in Table~\ref{tab:downstream} (see Appendix~\ref{app:downstream} for full details), demonstrate that \emph{HyPO} maintains or improves performance across all evaluated tasks relative to the corresponding baselines. This indicates that our method avoids additional performance degradation on downstream tasks that can be induced by alignment.

\textbf{Ablations and threshold $\gamma$ sensitivity.}
We ablate two enhancements in our \emph{HyPO} implementation: the use of a better reference model and a home advantage margin. The results in Table~\ref{tab:ablation-study} show that both components are beneficial. Notably, the ablated \emph{HyPO} variants still significantly outperform the standard DPO baseline, a finding consistent with our head-to-head comparisons in Figure~\ref{figure3}, confirming the effectiveness of our core mechanism. Furthermore, we analyze the model's sensitivity to the threshold $\gamma$ from Eq.~\ref{eq:hypo-margin-hard}. As shown in Figure~\ref{fig:your-plot}, performance remains stable across a range of $\gamma$ values around $0$. Given this low sensitivity, we adopt the most principled and interpretable setting, $\gamma=0$. These validate the contributions of our proposed method and the stability of our configuration.

\newlength{\abltabwidth}
\setlength{\abltabwidth}{0.54\textwidth}
\begin{figure*}[t]
\centering
\begin{minipage}[t]{\abltabwidth}
\centering
\vspace{0pt}
\small
\renewcommand{\arraystretch}{1.16}
\setlength{\tabcolsep}{2.0pt}
\captionof{table}{Ablation study of \emph{HyPO}'s components.}
\label{tab:ablation-study}
\vspace{-0.8em}
\resizebox{\abltabwidth}{!}{
\begin{tabular}{l|cc|cc|c}
\toprule
\emph{Llama-3-Models}  & \multicolumn{2}{c}{\textbf{Base}} & \multicolumn{2}{|c|}{\textbf{Instruct}} &\\ 
\midrule
& {\scriptsize LC(\%)} &  {\scriptsize WR(\%)}  & {\scriptsize LC(\%)}  & {\scriptsize WR(\%)} & {\scriptsize Avg. $\Delta$(\%)} \\
\midrule
HyPO (Ours) & 34.7  & 33.6 & 49.5 & 46.2 & \multicolumn{1}{c}{-}\\
$\quad$ - \emph{Home Advantage} & 33.2 & 32.5 & 47.4 & 45.1 & -\,1.5  \\
$\quad$ - \emph{Better Reference} & 29.3 & 27.1 & 45.3 & 45.9 & -\,4.1 \\
$\quad$ - \emph{\textcolor{changecolor}{BR and HA}} & \textcolor{changecolor}{28.8} & \textcolor{changecolor}{28.5} & \textcolor{changecolor}{45.2} & \textcolor{changecolor}{44.9} & -\,\textcolor{changecolor}{4.2}  \\
\midrule
Standard DPO & 24.3 & 21.9 & 40.9 & 41.3 & -\,8.9 \\
\bottomrule
\end{tabular}
}
\vspace{-0.5em}
\end{minipage}
\hfill
\begin{minipage}[t]{\dimexpr\textwidth-\abltabwidth-1em\relax}
\centering
\small
\captionof{figure}{\textcolor{changecolor}{Sensitivity to the threshold $\gamma$, using Meta-Llama-3-8B-Instruct.}}
\vspace{-0.8em}
\label{fig:your-plot}
\includegraphics[width=0.99\linewidth]{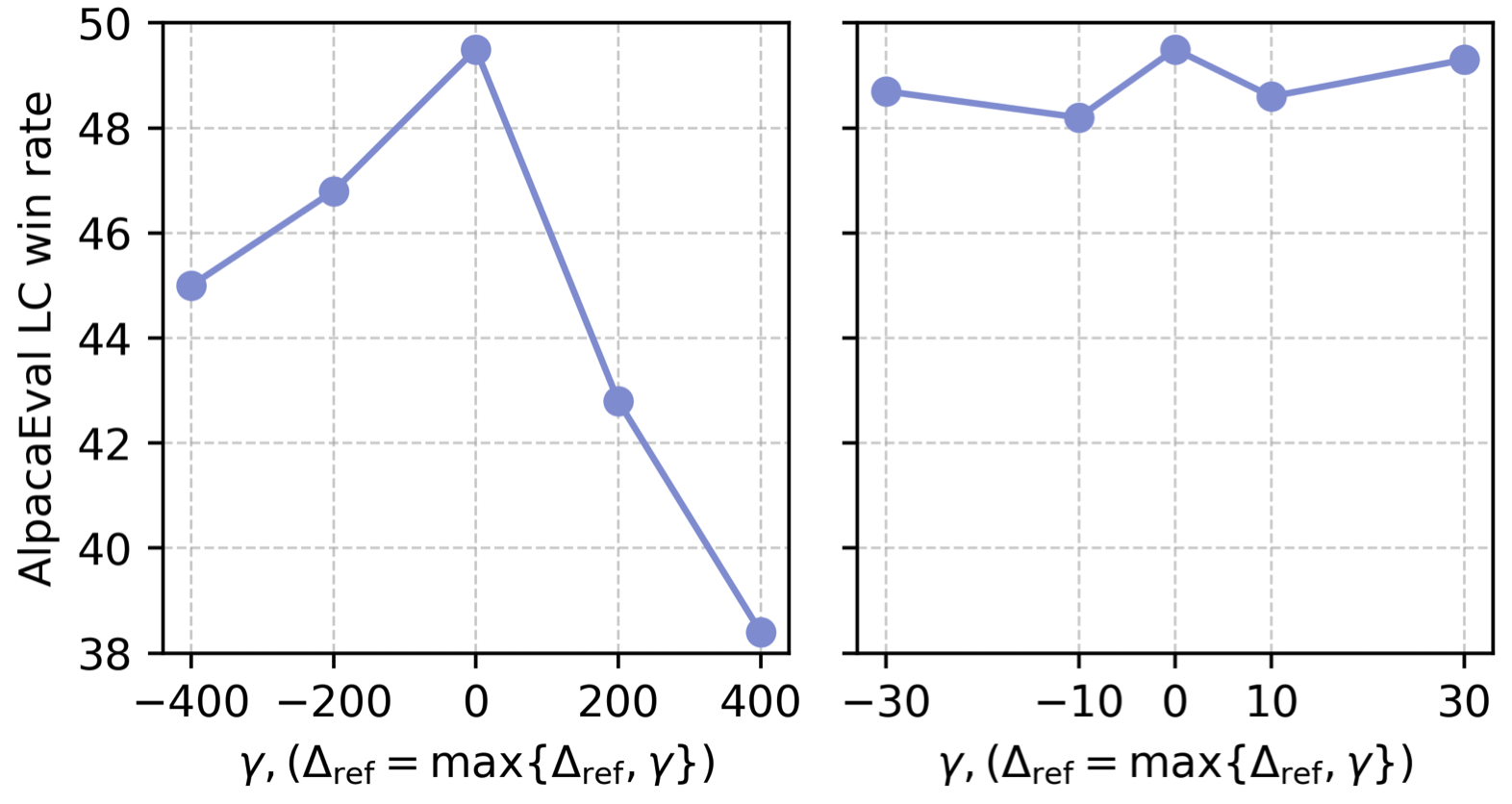}
\end{minipage}
\end{figure*}

\textbf{Running time.}  As a drop-in replacement for standard DPO, \emph{HyPO} introduces negligible additional computation. In the experiment shown in Figure~\ref{figure3}, conducted with identical hardware and settings, the wall clock training time for \emph{HyPO} differed from that of DPO by $\approx1\%$ (7.1 hours vs. 7.2 hours).

\textbf{Example Responses.}
To offer a qualitative view, Appendix~\ref{app:examples} compares sample model outputs. The responses suggest that \emph{HyPO} can generate more considered answers that acknowledge subtleties in the prompt, distinguishing it from the more direct outputs of baseline methods.

\textbf{Scalability.} To further validate the scalability of our proposed method, we conduct extended experiments across varying scales, datasets, and architectures. We evaluate \emph{HyPO} along three axes: (i) a dataset shift from UltraFeedback to HelpSteer2 \citep{wang2024helpsteer}; (ii) a model-size shift to Mistral-Nemo-12B \citep{nvidia-mistral-nemo-12b-instruct-2024}; and (iii) an architectural shift to Qwen-2.5-7B \citep{qwen25}. We follow the evaluation protocol and most hyperparameters from Section~\ref{settings} (details in Appendix~\ref{appendix_settings}). As summarized in Table~\ref{tab:extend_res}, \emph{HyPO} consistently improves over baselines across all settings, indicating robustness to distribution changes, parameter scaling, and diverse model families.

\begin{table*}[t]
\centering
\renewcommand{\arraystretch}{1.10}
\setlength{\tabcolsep}{1.8pt}
\small 
\caption{Further results reported from AlpacaEval 2.0 \citep{alpaca_eval, dubois2024length} and Arena-Hard-v0.1 \citep{li2024crowdsourced, arenahard2024}. LC and WR denote \emph{length-controlled win rate} and \emph{raw win rate}, respectively. The best results are highlighted in bold.}
\vspace{-0.5em}
\resizebox{\textwidth}{!}{
\begin{tabular}{lccccccccc}
\toprule
\multirow{3}{*}{\textbf{Method}} & \multicolumn{3}{c}{\scriptsize \textbf{Helpsteer2}} & \multicolumn{3}{c}{\scriptsize \textbf{Mistral-Nemo-Instruct (12B)}} & \multicolumn{3}{c}{\scriptsize \textcolor{changecolor}{\textbf{Qwen-2.5-7B}}} \\ 
\cmidrule(lr){2-4}\cmidrule(lr){5-7}\cmidrule(lr){8-10}
& \multicolumn{2}{c}{\scriptsize{\textbf{AlpacaEval}}} & \multicolumn{1}{c}{\scriptsize{\textbf{Arena-Hard}}}  & \multicolumn{2}{c}{\scriptsize{\textbf{AlpacaEval}}} & \multicolumn{1}{c}{\scriptsize{\textbf{Arena-Hard}}} & \multicolumn{2}{c}{\scriptsize{\textbf{\textcolor{changecolor}{AlpacaEval}}}} & \multicolumn{1}{c}{\scriptsize{\textbf{\textcolor{changecolor}{Arena-Hard}}}}  \\
\cmidrule(lr){2-3}\cmidrule(lr){4-4} \cmidrule(lr){5-6}\cmidrule(lr){7-7} \cmidrule(lr){8-9}\cmidrule(lr){10-10}
& {\scriptsize \bf LC (\%)} & {\scriptsize \bf WR (\%)}  & {\scriptsize \bf WR (\%)} & {\scriptsize \bf LC (\%)}  & {\scriptsize \bf WR (\%)}  & {\scriptsize \bf WR (\%)} & {\scriptsize \bf \textcolor{changecolor}{LC (\%)}}  & {\scriptsize \bf \textcolor{changecolor}{WR (\%)}}  & {\scriptsize \bf \textcolor{changecolor}{WR (\%)}} \\
\midrule
DPO~\citep{rafailov2023direct} &  16.2 & 14.3	 & 4.5 & 50.4 & 49.2 & 35.5  & \textcolor{changecolor}{27.9} & \textcolor{changecolor}{22.4} & \textcolor{changecolor}{28.8}   \\
SimPO~\citep{meng2024simpo} &  18.6 & 16.0 & 7.0 & 52.1 & 46.4 & 33.9 & \textcolor{changecolor}{33.7} & \textcolor{changecolor}{22.2} & \textcolor{changecolor}{33.8}  \\
\midrule
\emph{HyPO}~(Ours) &  \textbf{22.3} & \textbf{19.5} & \textbf{9.3} & \textbf{55.7} & \textbf{54.9} & \textbf{38.9} & \textbf{\textcolor{changecolor}{38.0}} & \textbf{\textcolor{changecolor}{30.7}} & \textbf{\textcolor{changecolor}{36.2}} \\
\bottomrule
\end{tabular}
}
\label{tab:extend_res}
\vspace{-0.6em}
\end{table*}

\section{Conclusion}
In this paper, we introduced \emph{Hybrid-DPO (HyPO)} to address \emph{premature satisfaction}, a critical failure in Direct Preference Optimization (DPO) where a pessimistic reference model down-weights learning signals. \emph{HyPO}'s simple conditional mechanism, clipping the reference margin at zero, corrects this bias while retaining DPO's inherent stability. Empirically, \emph{HyPO} consistently and significantly outperforms both various DPO and strong reference-free baselines across a range of models and benchmarks, proving robust and scalable without an additional ``alignment tax''. Our work provides evidence that direct preference alignment could be enhanced by conditionally debiasing the reference signal, rather than discarding it, thereby achieving a more effective and stable compromise.

\section{\textcolor{changecolor}{Limitations and Future Work}}
\label{sec:limitations}

\textcolor{changecolor}{Our approach aims to \emph{mitigate} rather than fully resolve the training-inference mismatch. By design, \emph{HyPO} retains the reference-based behavior on optimistic pairs to preserve training stability and leverage better reference models, meaning the training–inference mismatch persists for this subset. Furthermore, our method assumes general label reliability; in cases of severe label noise where the reference model correctly disagrees with a wrong label, \emph{HyPO} enforces a strong learning signal and may arguably amplify noise compared to DPO. Finally, while we prioritized a minimalist, parameter-free clipping mechanism, more sophisticated functional forms, such as adaptive clipping thresholds or learned gating mechanisms, remain promising directions for future research.}

\subsection*{Acknowledgments}
The authors would like to thank the anonymous reviewers for their insightful and constructive comments.
Suqin Yuan thanks Muyang Li and Li He for their valuable advice.
Tongliang Liu is partially supported by the following Australian Research Council projects: FT220100318, DP260102466, DP220102121, LP220100527, LP220200949.
Suqin Yuan is supported in part by the OpenAI Researcher Access Program (projects 05950 and 19000).
This research/project is supported by the National Research Foundation, Singapore under its National Large Language Models Funding Initiative (AISG Award No: AISG-NMLP-2024-003). Any opinions, findings and conclusions or recommendations expressed in this material are those of the author(s) and do not reflect the views of National Research Foundation, Singapore. 

\subsection*{Ethics statement}
Our research aims to improve the alignment of large language models with human preferences, a goal intended to enhance their safety, helpfulness, and reliability. We primarily use publicly available datasets, such as UltraFeedback \citep{cui2023ultrafeedback} and HelpSteer2 \citep{wang2024helpsteer}, which are aggregated from existing sources. While we did not collect new human data, we acknowledge that these datasets may contain inherent biases or sensitive information reflective of their origins. Our work does not aim to create new models from scratch but to refine existing ones (Mistral and Llama-3; \citealp{jiang2023clip, llama3modelcard}). We recognize that any advancement in AI capabilities carries a dual-use risk \citep{hong2024improving, huang2024machine, lin2025force, zhangstair, huang2026gradient}. However, the explicit goal of our method, \emph{HyPO}, is to make models' behavior more predictable and aligned with positive human values, thereby mitigating risks associated with misaligned AI. We believe that research into robust alignment techniques is a crucial step toward the responsible development and deployment of AI systems.

\subsection*{Reproducibility statement}
\label{appendix_reproducibility}
Our method, Hybrid-DPO (HyPO), is designed as a minimalist, plug-in modification to the standard Direct Preference Optimization (DPO) objective. This design ensures that our work can be easily reproduced by integrating our logic into popular alignment libraries and codebases that already support DPO, such as Alignment-handbook \citep{Tunstall_The_Alignment_Handbook}. 
All models and datasets used in our experiments are open-source and publicly accessible through the Hugging Face, ensuring full transparency. The base models include Mistral-7B \citep{jiang2023clip}, Meta-Llama-3-8B \citep{llama3modelcard}, and Mistral-Nemo-12B \citep{nvidia-mistral-nemo-12b-instruct-2024}. The datasets for supervised fine-tuning and preference alignment include UltraChat-200k \citep{ding2023enhancing}, UltraFeedback \citep{cui2023ultrafeedback}, HelpSteer2 \citep{wang2024helpsteer}, and OpenAssistant2 \citep{kopf2023openassistant}. For evaluation on benchmarks like AlpacaEval 2.0 \citep{alpaca_eval, dubois2024length}, Arena-Hard-v0.1 \citep{li2024crowdsourced, arenahard2024}, and LM Evaluation Harness \citep{biderman2024lessons}, we followed established protocols that utilize the GPT-4-Preview-1106 API as an automated judge. While access to this commercial API is managed by a third party, our evaluation methodology is public and can be adapted for use with other capable judge models. We have release our code to facilitate further research: \url{https://github.com/tmllab/2026_ICLR_HyPO}.

\subsection*{The Use of Large Language Models} 
During the preparation of this manuscript, the authors utilized large language models (LLMs) in several capacities. As writing and coding assistants, they were used to improve grammar, spelling, and sentence structure for clarity and to generate code for training and visualization. The subject of this research is the alignment of large language models themselves, by using LLMs such as Mistral and Llama-3 \citep{jiang2023clip, llama3modelcard}. Furthermore, as part of the evaluation methodology, the authors employed a closed-source LLM's API (GPT-4-Preview-1106) to act as an automated judge for assessing the quality of the trained LLMs.
The authors maintained full intellectual control of this paper. LLMs were used as tools to augment the research process, and the final manuscript reflects the authors' own work and insights.

\bibliography{iclr2025_conference}
\bibliographystyle{iclr2025_conference}

\appendix

\clearpage

\section{Experimental Details}
\label{appendix_settings}

This section provides a comprehensive overview of the experimental settings used to validate our proposed method, \emph{Hybrid-DPO (HyPO)}. Our experimental design is grounded in established practices from recent alignment literature to ensure fair, robust, and reproducible comparisons \citep{tunstall2023zephyr, meng2024simpo, Tunstall_The_Alignment_Handbook}. We detail the models and datasets, the multi-stage training pipeline, hyperparameter tuning, and the evaluation protocols for all experiments.

\subsection{Models and Datasets}

Our experiments leverage a suite of publicly available models and datasets to ensure full transparency and facilitate replication.
We conduct experiments on two popular open-source model families: Mistral-7B \citep{jiang2023clip} and Meta-Llama-3-8B \citep{llama3modelcard}. We test each model in two settings: a \emph{Base} setting, starting from the raw pretrained weights, and an \emph{Instruct} setting, starting from the official instruction-tuned checkpoints. For our scalability study, we also use Mistral-Nemo-Instruct-12B \citep{nvidia-mistral-nemo-12b-instruct-2024}.

\textbf{Supervised Fine-Tuning (SFT) Datasets.} For experiments in the \emph{Base} setting, we first conduct supervised fine-tuning to obtain a capable initial policy. We use UltraChat-200k \citep{ding2023enhancing} for the main experiments and OpenAssistant2 \citep{kopf2023openassistant} for the dataset transfer experiment.

\textbf{Preference Datasets.} For the core preference alignment stage, our main experiments use the UltraFeedback dataset \citep{cui2023ultrafeedback}, from which we use the official training split for optimization and hold out a fixed validation set. For the dataset transfer scalability experiment, we use the HelpSteer2 dataset \citep{wang2024helpsteer}. In \emph{Instruct} setting, we use UltraFeedback dataset released by \cite{meng2024simpo}.

\subsection{Pipeline and Hyperparameters}

Our training process consists of two main stages for the \emph{Base} setting (SFT followed by preference alignment) and one stage for the \emph{Instruct} setting (preference alignment only).

\textbf{Supervised Fine-Tuning (SFT) Stage.}

For the \emph{Base} models, we first create an SFT version by training on the relevant dataset for one epoch. We use the AdamW optimizer \citep{loshchilov2017decoupled} with a global batch size of 128, a peak learning rate of $2 \times 10^{-5}$, and a cosine learning rate schedule with a 10\% warm-up phase. The maximum sequence length is set to 2048. This SFT model serves as both the initial policy and the default reference model ($\pi_{\text{ref}}$) for preference alignment.

\textbf{Preference Alignment Stage.}

We conduct a small targeted hyperparameter search for each method.

\emph{\textcolor{changecolor}{General Hyperparameters.}} 
\textcolor{changecolor}{To ensure consistent optimization dynamics, across all preference alignment experiments, we train for a single epoch with a global batch size of 128, a maximum sequence length of 2048, and a maximum prompt length of 1800 tokens. 
We use the AdamW optimizer \citep{loshchilov2017decoupled} with $\beta_1=0.9$ and $\beta_2=0.999$, and a cosine learning rate schedule with a 10\% warm-up phase. 
To maintain the fixed global batch size of 128 despite varying memory requirements across stages (e.g., the additional reference model overhead in DPO) and hardware configurations (ranging from 1 to 4 GPUs), we adjusted the per-device micro-batch size and gradient accumulation steps based on the available GPU memory and model scale. For any specific set of comparative experiments, we ensured that these effective batch size settings were strictly identical across all baselines to guarantee a fair comparison.}

\emph{Method-Specific Hyperparameters.} We identified the peak learning rate and the DPO temperature $\beta$ as the most sensitive hyperparameters. For each method we reproduced, we performed a small grid search and selected the best configuration based on the lowest validation loss on our held-out set. In our experiments, we used some of the checkpoints released by \cite{meng2024simpo}.
\begin{itemize}[leftmargin=0.4cm,topsep=-2pt]
\item [-]
\textit{Peak Learning Rate:} Searched over $\{5 \times 10^{-7}, 1 \times 10^{-6}\}$.
\item [-]
\textit{Temperature $\beta$}: For standard objectives (DPO, CPO, KTO, FocalPO, TR-DPO, and HyPO.), we searched over $\{0.01, 0.1\}$. For methods employing length normalization (e.g., RainbowPO, SimPO), we followed \cite{meng2024simpo} recommendations and searched over $\{2, 10\}$.
\end{itemize}

\textbf{Evaluation Stage.}

\emph{Instruction-Following Benchmarks.}
We use GPT-4-Preview-1106 as the automated judge. The primary metric is the win rate (\%). For AlpacaEval 2.0 \citep{alpaca_eval, dubois2024length}, we report both the raw win rate (WR) and the length-controlled win rate (LC WR) against the GPT-4-Preview-1106 baseline model. Following \cite{meng2024simpo, tunstall2023zephyr}, we employed a sampling-based decoding approach. The temperature was set to 0.7 for the Mistral-Base \citep{jiang2023clip} setting, 0.5 for the Mistral-Instruct models (7B; \citealp{jiang2023clip}/12B; \citealp{nvidia-mistral-nemo-12b-instruct-2024}), and 0.9 across both Llama-3-8B \citep{llama3modelcard}. For Arena-Hard-v0.1 \citep{li2024crowdsourced, arenahard2024}, we report the win rate against the GPT-4-0314 baseline model, and we use greedy decoding.

\emph{Downstream Task Evaluation.}
To assess whether alignment methods introduce an \emph{alignment tax} by degrading general capabilities, we use the LM Evaluation Harness \citep{biderman2024lessons}. We evaluate performance on MMLU (5-shot) \citep{hendrycks2020measuring}, ARC-Challenge (25-shot) \citep{clark2018think}, HellaSwag (10-shot) \citep{zellers2019hellaswag}, Winogrande (5-shot) \citep{sakaguchi2021winogrande}, TruthfulQA (0-shot) \citep{lin2021truthfulqa}, and GSM8K (5-shot) \citep{cobbe2021training}.

\subsection{\textcolor{changecolor}{Implementation and infrastructure details}}
\label{app:impl_details}

\textcolor{changecolor}{
Our implementation builds on the open-source Alignment-handbook codebase~\citep{Tunstall_The_Alignment_Handbook}, which itself is built around Hugging Face Transformers and the \texttt{trl} library. 
HyPO is realized as a thin modification of \texttt{trl}'s \texttt{DPOTrainer} (see Appendix~G); all remaining components (data loading, logging, checkpointing, evaluation scripts) are reused from the Alignment-handbook stack with minimal changes. See Appendix~\ref{app:impl_details} for low-level implementation details.}

\textcolor{changecolor}{
We train all models with DeepSpeed \citep{rasley2020deepspeed} ZeRO and bfloat16 mixed precision. 
For the main runs, we use ZeRO stage-1 with CPU offloading of optimizer states, which substantially reduces GPU memory usage while keeping the implementation simple.
On a subset of configurations, we also ran ZeRO stage-3 without CPU offloading and observed that all reported metrics (loss, win-rate) match well. 
Based on this observation, we adopted the stage-1 + CPU-offload configuration for all large-scale sweeps reported in the paper.
All main results are obtained on compute nodes equipped with four NVIDIA H100 96GB GPUs.
We have verified that the same configuration can be run on two H100 GPUs, or on a single H100 with CPU offloading, at the cost of longer wall-clock time but essentially identical performance
Gradient checkpointing is enabled for all models to further reduce memory consumption.
Unless otherwise stated, we use Flash Attention 2 \citep{dao2023flashattention} for faster attention computation, bfloat16 mixed-precision for accelerated throughput, and realized via data parallelism and gradient accumulation.}

\subsection{Analysis of Reference Model Pessimism (Figure 2)}
To motivate our work, the analysis in Figure \ref{figure2} quantifies the prevalence of pessimistic reference margins ($\Delta_{\mathrm{ref}} < 0$). This experiment was conducted on the full training split of the UltraFeedback dataset \citep{cui2023ultrafeedback}. We computed the sequence-level log-likelihood margins for three distinct reference models derived from the Meta-Llama-3-8B family: (i) the raw base model, (ii) our SFT checkpoint trained on UltraChat-200k (use Zephyr settings; \citealp{tunstall2023zephyr}), and (iii) a model further aligned from the SFT checkpoint using SimPO (use SimPO settings; \citealp{meng2024simpo}). For each pair in the dataset, we calculated $\Delta_{\mathrm{ref}}$  without length normalization. We provide summary statistics for each distribution, including the mean and median (p50). The results confirmed that even for a strong, pre-aligned reference model, a substantial fraction of pairs ($\approx45\%$) remained in the pessimistic region, underscoring the persistence of the training-inference mismatch.

\subsection{Direct DPO vs. HyPO Comparison (Figure 3)}
Figure~\ref{figure3} presents a controlled comparison between standard DPO and our proposed \emph{HyPO}. The goal of this experiment was to isolate the impact of our conditional reference mechanism. Both methods started from the exact same SFT checkpoint of Llama-3-8B-Base and were trained on UltraFeedback \citep{cui2023ultrafeedback}. Critically, all hyperparameters were held identical and were based on the DPO configuration from Zephyr \citep{tunstall2023zephyr}, including a learning rate of $5 \times 10^{-7}$ and $\beta=0.01$. For this specific comparison, \emph{HyPO} was used in its most basic form: it used the SFT model as its reference (not a better reference) and was configured with $\gamma=0$ and no home advantage margin ($h=0$). This setup ensures a fair, one-to-one comparison against the DPO baseline. The metrics in Figure \ref{figure3}(a) and \ref{figure3}(b) were computed periodically on our held-out validation set throughout the single training epoch. The win-rate matrix in Figure \ref{figure3}(c) was generated by evaluating the final SFT, DPO, and HyPO checkpoints against each other on AlpacaEval \citep{alpaca_eval}.

\section{\textcolor{changecolor}{Analysis of DPO with SFT Loss}}
\label{app:dpo_sft_baseline}

\textcolor{changecolor}{In this section, we explore a natural baseline: adding a supervised fine-tuning (SFT) loss on the chosen response to the standard DPO objective. This acts as a regularizer to ensure the model maintains high likelihood for preferred responses. The $\text{DPO + SFT-on-chosen}$ objective is defined as:}
\begin{equation}
\textcolor{changecolor}{\mathcal{L}_{\text{DPO + SFT-on-chosen}} = \mathcal{L}_{\text{DPO}} + \lambda \, \mathcal{L}_{\text{SFT}}(y_w),}
\end{equation}
\textcolor{changecolor}{where $\mathcal{L}_{\text{SFT}}(y_w) = -\log \pi_\theta(y_w \mid x)$.}

\subsection{\textcolor{changecolor}{Hyperparameter Search}}
\textcolor{changecolor}{To ensure a fair comparison, we conducted a hyperparameter sweep for the coefficient $\lambda$ over the range $\{0.01, 0.03, 0.1, 0.3, 0.5\}$. 
We established $\lambda=0.5$ as the upper bound for our sweep because we observed that, at this value, the magnitude of the SFT loss term was approximately equal to the DPO loss term at the beginning of training. We posit that the auxiliary SFT loss should act as a regularizer rather than the dominant training signal; therefore, $\lambda$ should not exceed the point where SFT dominates the preference optimization.
Empirically, we found that the optimal performance on AlpacaEval 2.0 was achieved at $\lambda=0.03$. We use this best-performing configuration for the definitive comparison below.}

\subsection{\textcolor{changecolor}{Comparison and Analysis}}
\textcolor{changecolor}{We compared the optimized $\text{DPO + SFT-on-chosen}$ ($\lambda=0.03$) against the standard DPO and our HyPO method. In addition to win rates, we measured the Agreement Rate (defined in Appendix~\ref{appednix_metric}) to assess how well the model's likelihood ordering aligns with the ground truth preferences.}

\begin{table}[h]
\centering
\caption{\textcolor{changecolor}{Comparison of DPO, DPO + SFT-on-chosen ($\lambda=0.03$), and HyPO. All models are Llama-3-Base-8B initialized from the same SFT checkpoint.}}
\label{tab:sft_comparison_main}
\vspace{0.2em}
\begin{tabular}{lccc}
\toprule
\textbf{Method} & \textbf{AlpacaEval LC (\%)} & \textbf{AlpacaEval WR (\%)} & \textbf{Agreement Rate (\%)} \\
\midrule
DPO (Baseline) & 24.3 & 21.9 & 55.9 \\
DPO + SFT-on-chosen & 26.1 & 21.9 & 56.3 \\
\textbf{HyPO (Ours)} & \textbf{28.8} & \textbf{28.5} & \textbf{60.4} \\
\bottomrule
\end{tabular}
\end{table}

\textcolor{changecolor}{As shown in Table~\ref{tab:sft_comparison_main}, $\text{DPO + SFT-on-chosen}$ ($\lambda=0.03$) yields only a marginal improvement in win rate (+1.8\% LC) and Agreement Rate (+0.4\%). HyPO significantly outperforms this baseline (+4.5\% LC over baseline).}

\clearpage

\section{Examples Responses}
\label{app:examples}

The following is an example question along with responses from various models from the Alpaca Eval 2.0 helpful subset base pipeline. HyPO response shows analytical thinking, a deeper reasoning process, and consideration of implicit factors.

\begin{itemize}[leftmargin=0.4cm,topsep=-2pt]
\item [1.]
gpt4-1106-preview: Most comprehensive and beginner-safe; actionable progressions. 
\item [2.]
 HyPO (ours): Well-structured with reps/warmup/cooldown; a few choices (Russian twists, Pallof press) may be advanced or equipment-dependent.
\item [3.]
SimPO: Solid variety and set guidance; minor glitch (stray char) and several advanced core picks for true beginners. 
\item [4.]
DPO: Long single list with multiple advanced moves (hanging leg raises); fewer regressions and weaker structure. 
\end{itemize}

\begin{tcolorbox}[
    enhanced,
    nobeforeafter,
    colback=white, 
    colframe=mydarkblue, 
    colbacktitle=mydarkblue, 
    coltitle=white, 
    title=\textbf{Responses from Different Models}, 
    fonttitle=\bfseries, 
    boxrule=1pt,
    arc=2mm, 
    toprule=0.7mm, 
    bottomrule=0.7mm, 
    leftrule=0.7mm, 
    rightrule=0.7mm, 
    titlerule=0mm, 
    boxsep=4pt,
    left=4pt, 
    right=4pt, 
    top=3pt, 
    bottom=2pt 
]
\begin{tcolorbox}[
    enhanced,
    colback=white,
    colframe=CadetBlue,
    colbacktitle=CadetBlue, 
    coltitle=white,
    title=\textbf{Question}, 
    fonttitle=\bfseries, 
    boxrule=1pt, 
    arc=2mm,
    toprule=0.7mm, 
    bottomrule=0.7mm, 
    leftrule=0.7mm, 
    rightrule=0.7mm, 
    titlerule=0mm, 
    boxsep=4pt, 
    left=4pt, 
    right=4pt, 
    top=3pt, 
    bottom=2pt 
]
What are the best exercises for beginners?
\end{tcolorbox}
\medskip
\textbf{GPT-4-Preview-1106 (Baseline):} For beginners, the best exercises are those that are easy to learn, safe, and effective at building strength, endurance, and flexibility. Here is a list of exercises that are generally well-suited for beginners, covering different aspects of fitness:

\begin{enumerate}[nosep]
    \item \textbf{Cardiovascular Exercises:}
    \begin{itemize}[nosep]
        \item Walking: Start with a brisk walk and gradually increase your pace and distance.
        \item Cycling: Begin with a flat route before tackling hills.
        \item Swimming: Try a few laps with rest in between, gradually increasing your endurance.
        \item Jumping Jacks: A simple full-body exercise to get your heart rate up.
    \end{itemize}

    \item \textbf{Strength Training Exercises:}
    \begin{itemize}[nosep]
        \item Bodyweight Squats: Focus on form and depth before adding weight.
        \item Push-ups: Start with wall or knee push-ups if a full push-up is too difficult.
        \item Planks: Begin by holding the position for short periods and gradually increase the time.
        \item Lunges: Practice stationary lunges to master the form before attempting walking lunges.
        \item Dumbbell Exercises: Use light weights to perform exercises like bicep curls, shoulder presses, and tricep extensions.
    \end{itemize}

    \item \textbf{Flexibility and Balance Exercises:}
    \begin{itemize}[nosep]
        \item Stretching: Incorporate basic stretches for all major muscle groups, holding each stretch for 15-30 seconds.
        \item Yoga: Basic poses like Child's Pose, Downward-Facing Dog, and Warrior can improve flexibility and balance.
        \item Tai Chi: This gentle martial art is great for balance and stress reduction.
    \end{itemize}

    \item \textbf{Core Strengthening Exercises:}
    \begin{itemize}[nosep]
        \item Bird-Dog: This exercise improves core stability and coordination.
        \item Dead Bug: Another great move for core strength that's easy on the back.
        \item Seated Russian Twists: Begin without weight and focus on the twisting motion.
    \end{itemize}

    \item \textbf{Functional Exercises:}
    \begin{itemize}[nosep]
        \item Step-ups: Use a low step at first and focus on form.
        \item Farmer's Walk: Carry a light weight in each hand and walk for a set distance, maintaining good posture.
        \item Glute Bridges: These help strengthen the glutes and stabilize the core.
    \end{itemize}
\end{enumerate}

It's essential for beginners to focus on proper form to prevent injuries and to start with a level of intensity that matches their current fitness level. It's also advisable to consult with a fitness professional or a physician before starting any new exercise program, especially if you have pre-existing health conditions or concerns.
\end{tcolorbox}

\begin{tcolorbox}[
    enhanced,
    colback=white, 
    colframe=mydarkblue, 
    colbacktitle=mydarkblue, 
    coltitle=white, 
    title=\textbf{Responses from Different Models}, 
    fonttitle=\bfseries, 
    boxrule=1pt,
    arc=2mm, 
    toprule=0.7mm, 
    bottomrule=0.7mm, 
    leftrule=0.7mm, 
    rightrule=0.7mm, 
    titlerule=0mm, 
    boxsep=4pt,
    left=4pt, 
    right=4pt, 
    top=3pt, 
    bottom=2pt 
]

\begin{tcolorbox}[
    enhanced,
    colback=white,
    colframe=CadetBlue,
    colbacktitle=CadetBlue, 
    coltitle=white,
    title=\textbf{Question}, 
    fonttitle=\bfseries, 
    boxrule=1pt, 
    arc=2mm,
    toprule=0.7mm, 
    bottomrule=0.7mm, 
    leftrule=0.7mm, 
    rightrule=0.7mm, 
    titlerule=0mm, 
    boxsep=4pt, 
    left=4pt, 
    right=4pt, 
    top=3pt, 
    bottom=2pt 
]
What are the best exercises for beginners?
\end{tcolorbox}
\medskip
\textbf{DPO:} As a beginner, it's essential to start with exercises that are safe, effective, and enjoyable to help you build a strong foundation and momentum in your fitness journey. Here are some of the best exercises for beginners:

\begin{enumerate}[nosep]
    \item \textbf{Squats:} Works your legs, glutes, and core. Stand with feet shoulder-width apart, then bend knees and lower your body down, keeping back straight.
    \item \textbf{Push-ups:} Targets chest, shoulders, and triceps. Start in a plank position with hands shoulder-width apart, lower your body until your chest almost touches the ground, then push back up.
    \item \textbf{Lunges:} Works legs, glutes, and lower body. Stand with feet together, take a large step forward with one foot, and lower your body down, keeping back knee almost touching the ground. Alternate legs.
    \item \textbf{Planks:} Strengthens core and improves posture. Hold a plank position for 30-60 seconds, engaging your abs and keeping your body straight from head to heels.
    \item \textbf{Chest Press (using dumbbells or resistance bands):} Works chest, shoulders, and triceps. Lie on a flat surface or sit with knees bent, holding weights or resistance bands, and press them upwards towards the ceiling.
    \item \textbf{Rows (using dumbbells or resistance bands):} Targets back and biceps. Hold weights or resistance bands and pull them towards your chest, keeping your elbows close to your body.
    \item \textbf{Bicycle Crunches:} Engages core and improves obliques. Lie on your back with hands behind your head, alternate bringing each elbow towards the opposite knee, as if pedaling a bicycle.
    \item \textbf{Leg Raises (hanging leg raises or leg lifts):} Works abs and lower back. Hang from a pull-up bar or lie on your back with arms extended, lifting your legs straight up towards the ceiling.
    \item \textbf{Wall Sit:} Strengthens legs and glutes. Stand with back against a wall, slide down into a seated position with knees bent at a 90-degree angle, and hold for 30 seconds to 1 minute.
    \item \textbf{Brisk Walking or Jogging:} Improves cardiovascular fitness and burns calories. Start with short sessions and gradually increase duration and intensity.
\end{enumerate}

Remember to:
\begin{itemize}[nosep]
    \item Start slow and focus on proper form and technique.
    \item Begin with lighter weights and gradually increase the load as you build strength.
    \item Warm up before exercising with 5-10 minutes of light cardio and stretching.
    \item Listen to your body and rest when needed.
    \item Consult with a healthcare professional or fitness expert if you have any concerns or injuries.
\end{itemize}

As you progress, you can incorporate more exercises and intensity to challenge yourself further. Happy exercising!
\end{tcolorbox}

\begin{tcolorbox}[
    enhanced,
    colback=white, 
    colframe=mydarkblue, 
    colbacktitle=mydarkblue, 
    coltitle=white, 
    title=\textbf{Responses from Different Models}, 
    fonttitle=\bfseries, 
    boxrule=1pt,
    arc=2mm, 
    toprule=0.7mm, 
    bottomrule=0.7mm, 
    leftrule=0.7mm, 
    rightrule=0.7mm, 
    titlerule=0mm, 
    boxsep=4pt,
    left=4pt, 
    right=4pt, 
    top=3pt, 
    bottom=2pt 
]
\begin{tcolorbox}[
    enhanced,
    colback=white,
    colframe=CadetBlue,
    colbacktitle=CadetBlue, 
    coltitle=white,
    title=\textbf{Question}, 
    fonttitle=\bfseries, 
    boxrule=1pt, 
    arc=2mm,
    toprule=0.7mm, 
    bottomrule=0.7mm, 
    leftrule=0.7mm, 
    rightrule=0.7mm, 
    titlerule=0mm, 
    boxsep=4pt, 
    left=4pt, 
    right=4pt, 
    top=3pt, 
    bottom=2pt 
]
What are the best exercises for beginners?
\end{tcolorbox}
\medskip
\textbf{SimPO:} As a beginner, it's essential to start with exercises that are safe, effective, and manageable to build a strong foundation for your fitness journey. Here are some of the best exercises for beginners:

\begin{enumerate}[nosep]
    \item \textbf{Bodyweight exercises:}
    \begin{itemize}[nosep]
        \item Push-ups: Works chest, shoulders, and triceps.
        \item Squats: Targets legs, glutes, and core.
        \item Lunges: Works legs, glutes, and improves balance.
        \item Planks: Engages core muscles for stability and endurance.
        \item Pull-ups (or assisted pull-ups): Targets back, shoulders, and arms.
    \end{itemize}
    
    \item \textbf{Resistance band exercises:}
    \begin{itemize}[nosep]
        \item Banded squats, lunges, and chest presses: Add variety to bodyweight exercises with resistance bands.
    \end{itemize}
    
    \item \textbf{Cardio exercises:}
    \begin{itemize}[nosep]
        \item Brisk walking: Easy to start and improves cardiovascular fitness.
        \item Jogging or running (begin with short intervals): Increases endurance and burns calories.
        \item Swimming or cycling: Low-impact alternatives for those with joint concerns.
    \end{itemize}
    
    \item \textbf{Upper body exercises with light weights or dumbbells:}
    \begin{itemize}[nosep]
        \item Dumbbell chest press: Targets chest and shoulders.
        \item Bicep curls: Works biceps.
        \item Tricep dips (using a chair or bench): Targets triceps.
    \end{itemize}
    
    \item \textbf{Core exercises:}
    \begin{itemize}[nosep]
        \item Russian twists: Engages obliques.
        \item Leg raises: Targets lower abs.
        \item Bicycle crunches: Works entire core.
    \end{itemize}
\end{enumerate}

Remember to:
\begin{enumerate}[nosep]
    \item Start slowly and gradually increase intensity and duration as you become more comfortable.
    \item Focus on proper form and technique to avoid injuries.
    \item Begin with 2-3 sets of 8-12 repetitions for each exercise.
    \item Listen to your body and rest when needed.
    \item Consult with a healthcare professional or fitness expert to create a personalized workout plan.
\end{enumerate}

These exercises will help you build a solid foundation in strength, flexibility, and cardiovascular endurance. As you progress, you can always add more variety and intensity to your routine.
\end{tcolorbox}

\begin{tcolorbox}[
    enhanced,
    colback=white, 
    colframe=mydarkblue, 
    colbacktitle=mydarkblue, 
    coltitle=white, 
    title=\textbf{Responses from Different Models}, 
    fonttitle=\bfseries, 
    boxrule=1pt,
    arc=2mm, 
    toprule=0.7mm, 
    bottomrule=0.7mm, 
    leftrule=0.7mm, 
    rightrule=0.7mm, 
    titlerule=0mm, 
    boxsep=4pt,
    left=4pt, 
    right=4pt, 
    top=3pt, 
    bottom=2pt 
]
\begin{tcolorbox}[
    enhanced,
    colback=white,
    colframe=CadetBlue,
    colbacktitle=CadetBlue, 
    coltitle=white,
    title=\textbf{Question}, 
    fonttitle=\bfseries, 
    boxrule=1pt, 
    arc=2mm,
    toprule=0.7mm, 
    bottomrule=0.7mm, 
    leftrule=0.7mm, 
    rightrule=0.7mm, 
    titlerule=0mm, 
    boxsep=4pt, 
    left=4pt, 
    right=4pt, 
    top=3pt, 
    bottom=2pt 
]
What are the best exercises for beginners?
\end{tcolorbox}
\medskip
\textbf{HyPO (ours):} As a beginner, it's essential to start with exercises that are gentle, easy to learn, and progressive in intensity to build a strong foundation and prevent injury. Here are some of the best exercises for beginners:

\begin{enumerate}[nosep] 
    \item \textbf{Bodyweight exercises:}
    \begin{itemize}[nosep]
        \item Push-ups
        \item Squats
        \item Lunges
        \item Planks
        \item Bicycle crunches (modified or regular)
        \item Wall sit
    \end{itemize}
    
    \item \textbf{Resistance band exercises:}
    \begin{itemize}[nosep]
        \item Banded squats and lunges
        \item Banded chest press (using a resistance band or light dumbbells)
        \item Banded rows (using a resistance band)
    \end{itemize}
    
    \item \textbf{Low-impact cardio:}
    \begin{itemize}[nosep]
        \item Brisk walking
        \item Jogging in place
        \item Swimming (if accessible)
        \item Cycling (stationary or outdoor)
    \end{itemize}
    
    \item \textbf{Upper body strength exercises with light weights:}
    \begin{itemize}[nosep]
        \item Dumbbell rows
        \item Shoulder presses (using light weights or resistance bands)
        \item Bicep curls
        \item Tricep dips (using a chair or bench)
    \end{itemize}
    
    \item \textbf{Core exercises:}
    \begin{itemize}[nosep]
        \item Leg raises (lying or hanging)
        \item Russian twists (using a medicine ball or weight plate)
        \item Pallof press (using a cable machine or resistance band)
    \end{itemize}
\end{enumerate}

Remember to:
\begin{enumerate}[nosep]
    \item Start slow and gentle, with 2-3 sets of 8-12 repetitions for each exercise.
    \item Focus on proper form and technique over the number of repetitions or weight used.
    \item Listen to your body and rest when needed.
    \item Gradually increase intensity and volume as you become more comfortable with the exercises.
    \item Incorporate exercises that work multiple muscle groups at once (e.g., squats engage legs, glutes, and core).
\end{enumerate}

Additionally, consider the following general tips:
\begin{itemize}[nosep]
    \item Consult with a healthcare professional or fitness expert to create a personalized workout plan.
    \item Warm up before exercising with 5-10 minutes of light cardio and dynamic stretching.
    \item Cool down afterwards with static stretches to help prevent soreness and improve flexibility.
\end{itemize}

By starting with these beginner-friendly exercises, you'll set yourself up for a strong foundation in fitness and enjoy a positive exercise experience.

\end{tcolorbox}

\clearpage

\section{Downstream Tasks}

\textcolor{changecolor}{One of the primary challenges in preference alignment is the potential for a model's general capabilities to degrade, a phenomenon often attributed to over-optimization on the alignment objective \citep{gao2023scaling, rafailov2024scaling}. To rigorously assess whether our approach \textcolor{changecolor}{introduces any additional} ``alignment tax'' and preserves broad utility, we evaluated aligned models on a comprehensive suite of downstream tasks. For comparability, we benchmark against one reference-based baseline (DPO; \citealp{rafailov2023direct}) and one reference-free baseline (SimPO; \citealp{meng2024simpo}). All three methods are evaluated starting from the same SFT checkpoint, with identical prompt formatting and decoding hyperparameters.}

\textcolor{changecolor}{The evaluation was conducted using the standard LM Evaluation Harness framework \citep{biderman2024lessons}, with inference performed via the Hugging Face Transformers library. We follow the default task implementations in LM Evaluation Harness and report accuracy (or exact-match for GSM8K) as provided by the framework. We measured performance across four key domains, using the following few-shot settings for each task:}
\begin{itemize}[leftmargin=0.6cm,topsep=-2pt,itemsep=0pt]
\item \emph{Knowledge \& Language Understanding:} MMLU \citep{hendrycks2020measuring} (5-shot) and ARC-Challenge \citep{clark2018think} (25-shot).
\item \emph{Commonsense Reasoning:} HellaSwag \citep{zellers2019hellaswag} (10-shot) and Winogrande \citep{sakaguchi2021winogrande} (5-shot).
\item \emph{Factuality:} TruthfulQA \citep{lin2021truthfulqa} (0-shot).
\item \emph{Mathematical Reasoning:} GSM8K \citep{cobbe2021training} (5-shot).
\end{itemize}
The results, presented in Table~\ref{tab:downstream}, demonstrate that \emph{HyPO} consistently performs on par with or better than the initial model and other alignment methods. Notably, \emph{HyPO} achieves the highest average score in three out of the four experimental settings. This indicates that our method successfully aligns with human preferences without incurring a significant penalty on the model's core capabilities.

\label{app:downstream}
\begin{table}[h]
\renewcommand{\arraystretch}{1.1}
\setlength{\tabcolsep}{3.0pt}
    \caption{Downstream task evaluation results, using LM Evaluation Harness \citep{biderman2024lessons}. The best results are highlighted in bold, and the second-best are underlined.}
    \label{tab:downstream}
    \vspace{-0.5em}
    \resizebox{\textwidth}{!}{\begin{tabular}{@{}lccccccc@{}}
    \toprule
                   & \textbf{MMLU} & \textbf{ARC } & \textbf{HellaSwag} & \textbf{TruthfulQA} & \textbf{Winogrande} & \textbf{GSM8K} & \textbf{Average} \\ \midrule\midrule
                  
                   \multicolumn{8}{c}{{\color[HTML]{222222} \textbf{Mistral-Base (7B)}}}                                                                            \\ \toprule
Initial Model   & 59.0            & 54.2            & 60.8 & 28.0               & 77.7                & 34.6             & 52.4           \\
\midrule
DPO~\citep{rafailov2023direct}   & 57.6            & 58.7            & 64.4                  & 40.3                  & 77.4                & 31.5             & \underline{55.0}           \\
SimPO~\citep{meng2024simpo} & 58.4            & 58.6            & 63.6                  & 33.1                  & 78.1                & 35.0             & 54.5           \\
\midrule
\emph{HyPO}~(Ours) & 58.5            & 60.2             & 63.5                  & 39.1                  & 78.9               & 31.8            & \textbf{55.3}           \\ \toprule
    \multicolumn{8}{c}{{\color[HTML]{222222} \textbf{Mistral-Instruct (7B) }}}                                                                                      \\ \toprule
Initial Model   & 59.2            & 58.7            & 66.1                  & 52.6                  & 78.1                & 44.1             & 59.8           \\
\midrule
DPO~\citep{rafailov2023direct}   & 58.8            & 60.9            & 67.6                 & 51.2                  & 78.3                & 41.6             & 59.7           \\
SimPO~\citep{meng2024simpo} & 58.1            & 62.5            & 67.5                  & 55.0                  & 77.4                & 39.8             & \underline{60.1}           \\
\midrule
\emph{HyPO}~(Ours) & 59.6            & 66.8            & 67.4                  & 54.0                  & 77.6                & 40.6             & \textbf{61.0}           \\  \toprule
\multicolumn{8}{c}{{\color[HTML]{222222} \textbf{Llama-3-Base (8B)}}}                                                                                                                               \\ \toprule
Initial Model   & 63.7            & 56.9            & 61.4                  & 31.5                  & 77.4                & 50.3             & 56.9           \\
\midrule
DPO~\citep{rafailov2023direct}   & 63.3            & 61.6           & 64.7                  & 37.9                  & 78.0                & 54.7             & \underline{60.0}           \\
SimPO~\citep{meng2024simpo} & 62.1            & 63.4            & 64.5                  & 38.0                  & 77.2                & 50.4             & 59.3           \\
\midrule
\emph{HyPO}~(Ours) & 63.6            & 65.4            & 64.3                  & 42.8                  & 78.9                & 49.7             & \textbf{60.8}           \\    \toprule
\multicolumn{8}{c}{{\color[HTML]{222222} \textbf{Llama-3-Instruct (8B) }}}                                                                                       \\ \toprule
Initial Model   & 65.8            & 56.4            & 59.0                  & 36.1                  & 77.3                & 74.8             & 61.6           \\
\midrule
DPO~\citep{rafailov2023direct}   & 65.6            & 61.3            & 60.6                  & 42.4                  & 76.4                & 73.5             & \textbf{63.3}           \\
SimPO~\citep{meng2024simpo} & 64.7            & 59.8            & 56.0                  & 43.8                  & 72.5                & 71.3             & 61.4           \\
\midrule
\emph{HyPO}~(Ours) & 65.0            & 64.4            & 58.7                  & 45.3                  & 73.3                & 70.3             & \underline{62.8}           \\ 
\bottomrule
    \end{tabular}}
\end{table}

\clearpage

\section{Evaluation Metric Definitions}
\label{appednix_metric}

To provide a fine-grained analysis of model performance during training, as shown in Figure~\ref{figure3}, we track two key inference-aligned metrics on a held-out evaluation set of preference pairs $(x, y_w, y_l)$, where $y_w \succ y_l$. These metrics are designed to directly quantify how effectively our proposed method, \emph{HyPO}, mitigates the training-inference mismatch inherent in DPO. Both metrics are computed using the policy and reference log-likelihood margins, $\Delta_\theta$ and $\Delta_{\mathrm{ref}}$, as formally defined in the main text in Equation~\ref{eq:margins}. We denote the policy margin at a given training checkpoint $t$ as $\Delta_\theta^{(t)}$.

\textbf{Absolute Agreement Rate.}
The Absolute Agreement Rate provides a global measure of how well the policy's log-likelihoods align with the ground-truth preferences. It is defined as the probability that the policy assigns a higher log-likelihood to the winning response $y_w$ than the losing response $y_l$:
$$\operatorname{Agree}@t := \Pr_{(x,y_w,y_l)} \left[ \Delta_\theta^{(t)}(x, y_w, y_l) > 0 \right].$$
At inference time, a model's performance depends on its \emph{absolute} log-likelihoods, as the reference model is not used. This metric therefore directly reflects the desired inference-time behavior. A higher agreement rate signifies that the policy is more accurately ranking responses in absolute terms, indicating a successful reduction of the training-inference mismatch.

\textbf{Absolute Margin on the Pessimistic Subset.}
This metric is a targeted diagnostic designed to measure performance on the specific subset of data where DPO is most prone to failure. We first define the \emph{pessimistic subset} as all evaluation pairs where the reference model incorrectly prefers the losing response:
$$\mathcal{D}_{\text{pessimistic}} := \{ (x, y_w, y_l) \mid \Delta_{\mathrm{ref}}(x, y_w, y_l) < 0 \}.$$
The metric is then the average policy margin, $\Delta_\theta$, computed exclusively over this subset:
$$\text{Margin}_{\text{pessimistic}}@t := \mathbb{E}_{(x,y_w,y_l) \in \mathcal{D}_{\text{pessimistic}}} \left[ \Delta_\theta^{(t)}(x, y_w, y_l) \right].$$
This metric directly probes the phenomenon of \emph{premature satisfaction}. For pairs in $\mathcal{D}_{\text{pessimistic}}$, DPO's learning signal weakens as soon as the policy margin $\Delta_\theta$ surpasses the negative reference margin $\Delta_{\mathrm{ref}}$, even if $\Delta_\theta$ itself is still negative. A strong and increasing positive value for this metric demonstrates that the policy is successfully applying corrective pressure and overcoming the misleading signal from the pessimistic reference, a core goal of \emph{HyPO}.

\section{\textcolor{changecolor}{Analysis of the Smoothness Parameter $\alpha$}}
\label{app:smooth_alpha}

\textcolor{changecolor}{In Eq.~\ref{eq:hypo-softplus} we introduced a smooth variant of the clipped reference margin using a softplus transformation,
with smoothness parameter $\alpha$:}
\[
\textcolor{changecolor}{\widetilde{\Delta}_{\mathrm{ref}}
=\gamma+\frac{1}{\alpha}\log\!\big(1+\exp(\alpha(\Delta_{\mathrm{ref}}-\gamma))\big),\quad \alpha>0,}
\]
\textcolor{changecolor}{where smaller values of $\alpha$ yield a smoother transition, while $\alpha \to \infty$ recovers the hard
$\max\{\Delta_{\mathrm{ref}},\gamma\}$.}

\textcolor{changecolor}{To assess the sensitivity of \emph{HyPO} to this smoothness parameter, we ran a small sweep over
$\alpha \in \{1, 10, 100\}$ using Meta-Llama-3-8B-Instruct aligned on UltraFeedback, evaluated on AlpacaEval 2.0.
All other settings are identical to those in our main experiments.}

\begin{table}[h]
\centering
\small
\caption{\textcolor{changecolor}{Sensitivity of \emph{HyPO} to the smoothness parameter $\alpha$ in Eq.~\ref{eq:hypo-softplus}
on Meta-Llama-3-8B-Instruct.}}
\label{tab:alpha_sweep}
\begin{tabular}{lccc}
\toprule
Method & $\alpha$ & AlpacaEval LC (\%) & AlpacaEval WR (\%) \\
\midrule
HyPO           & 1   & 48.3 & 45.5 \\
HyPO (default) & 10  & 49.5 & 46.2 \\
HyPO           & 100 & 49.6 & 46.2 \\
\bottomrule
\end{tabular}
\end{table}

\textcolor{changecolor}{The results in Table~\ref{tab:alpha_sweep} show that \emph{HyPO} is robust to the choice of $\alpha$ in a reasonable range:
$\alpha=10$ and $\alpha=100$ yield nearly identical performance, while a much smoother transition ($\alpha=1$) causes only a modest degradation.
This supports our default choice of $\alpha=10$ as a simple and effective setting, and suggests that the benefits of \emph{HyPO}
do not rely on fine-tuning this parameter.}

\clearpage

\section{\textcolor{changecolor}{Minimal code change from DPO to HyPO}}

\textcolor{changecolor}{We implement HyPO by modifying only the computation of the DPO logits inside the original \texttt{DPOTrainer.dpo\_loss} in \texttt{trl}. 
Let}
\[
\textcolor{changecolor}{\Delta_\theta = \log \pi_\theta(y_{\text{c}} \mid x) - \log \pi_\theta(y_{\text{r}} \mid x),\qquad
\Delta_{\text{ref}} = \log \pi_{\text{ref}}(y_{\text{c}} \mid x) - \log \pi_{\text{ref}}(y_{\text{r}} \mid x),}
\]
\textcolor{changecolor}{be the policy and reference log-ratios. 
Standard DPO uses $\Delta_\theta - \Delta_{\text{ref}}$ as the logit input to the sigmoid loss. 
HyPO simply replaces $\Delta_{\text{ref}}$ with a clipped (or smoothed) reference margin $\widetilde{\Delta}_{\text{ref}}$:}
\[
\textcolor{changecolor}{\widetilde{\Delta}_{\text{ref}} = \max(\Delta_{\text{ref}}, \gamma)
\quad\text{or}\quad
\widetilde{\Delta}_{\text{ref}} = \gamma + \tau\,\mathrm{softplus}\!\left(\frac{\Delta_{\text{ref}}-\gamma}{\tau}\right),}
\]
\textcolor{changecolor}{where $\tau = 1/\alpha$, and uses $\Delta_\theta - \widetilde{\Delta}_{\text{ref}}$ in the DPO loss.}

\textcolor{changecolor}{The concrete code change in \texttt{DPOTrainer.dpo\_loss} is:}

\begin{lstlisting}[language=Python]
# Original DPO (simplified):
pi_logratios  = policy_chosen_logps - policy_rejected_logps
if self.reference_free:
    ref_logratios = torch.zeros_like(pi_logratios)
else:
    ref_logratios = reference_chosen_logps - reference_rejected_logps

logits = pi_logratios - ref_logratios
\end{lstlisting}

\textcolor{changecolor}{\noindent is replaced by the following HyPO version:}

\begin{lstlisting}[language=Python]
# HyPO: Conditional Reference Clipping
pi_logratios  = policy_chosen_logps - policy_rejected_logps
if self.reference_free:
    ref_logratios = torch.zeros_like(pi_logratios)
else:
    ref_logratios = reference_chosen_logps - reference_rejected_logps

# ---------------- HyPO Modification Start ----------------
gamma = torch.tensor(self.args.hypo_gamma, device=ref_logratios.device)

if getattr(self.args, "hypo_tau", 0.0) > 0.0:
    # Smooth HyPO: Eq. 13 (using tau for temperature)
    tau = self.args.hypo_tau
    ref_logratios = gamma + tau * F.softplus((ref_logratios - gamma) / tau)
else:
    # Hard HyPO: Eq. 11 (standard max)
    ref_logratios = torch.maximum(ref_logratios, gamma)
# ----------------- HyPO Modification End -----------------

logits = pi_logratios - ref_logratios
\end{lstlisting}

\textcolor{changecolor}{ 
In our experiments, we introduce \textit{hypo\_gamma} (scalar threshold $\gamma$, default 0) and optionally \textit{hypo\_tau} (smoothing factor $\tau = 1/\alpha$) as hyperparameters.}

\end{document}

%% file: math_commands.tex
\usepackage{amsmath,amsfonts,bm}

\def\eqref#1{equation~\ref{#1}}

\def\1{\bm{1}}

\DeclareMathAlphabet{\mathsfit}{\encodingdefault}{\sfdefault}{m}{sl}
\SetMathAlphabet{\mathsfit}{bold}{\encodingdefault}{\sfdefault}{bx}{n}

